\documentclass[12pt,preprint]{elsarticle} 

\usepackage{amssymb,amsthm,amsfonts,latexsym,cancel}
\usepackage[intlimits]{amsmath}
\usepackage{amsfonts}
\usepackage{latexsym}
\usepackage{titlesec}
\usepackage[english]{babel}
\usepackage{epstopdf}
\usepackage{epsfig}
\usepackage{caption}
\usepackage[pdftex]{hyperref}
\usepackage{longtable,multirow,booktabs}
\usepackage{stmaryrd}
\allowdisplaybreaks
\usepackage{graphicx} 
\usepackage{lscape}   

\usepackage{dsfont}
\usepackage{mathrsfs}
\usepackage{eucal} 

\usepackage{geometry}
\usepackage{todonotes} 
\usepackage{fullpage}
\usepackage{dirtree} 
\usepackage[inline]{trackchanges}
\addeditor{OLCG}
\addeditor{VD}
\addeditor{BG}

\begin{document}
\begin{frontmatter}

\title{Enhanced Vascular Flow Simulations in Aortic Aneurysm via Physics-Informed Neural Networks and Deep Operator Networks}

\author[label1]{Oscar L. Cruz-Gonz\'{a}lez} 
\author[label1]{Val\'{e}rie Deplano\corref{cor1}}\ead{valerie.deplano@univ-amu.fr}
\author[label2]{Badih Ghattas}

\address[label1]{Aix Marseille Univ, CNRS, Centrale Marseille, IRPHE UMR 7342, Marseille, France}
\address[label2]{Aix Marseille Univ, CNRS, AMSE UMR 7316, Marseille, France}\cortext[cor1]{Corresponding author}


\begin{abstract}

Due to the limited accuracy of 4D Magnetic Resonance Imaging (MRI) in identifying hemodynamics in cardiovascular diseases, the challenges in obtaining patient-specific flow boundary conditions, and the computationally demanding and time-consuming nature of Computational Fluid Dynamics (CFD) simulations, it is crucial to explore new data assimilation algorithms that offer possible alternatives to these limitations. In the present work, we study Physics-Informed Neural Networks (PINNs), Deep Operator Networks (DeepONets), and their Physics-Informed extensions (PI-DeepONets) in predicting vascular flow simulations in the context of a 3D Abdominal Aortic Aneurysm (AAA) idealized model. PINN is a technique that combines deep neural networks with the fundamental principles of physics, incorporating the physics laws, which are given as partial differential equations, directly into loss functions used during the training process. On the other hand, DeepONet is designed to learn nonlinear operators from data and is particularly useful in studying parametric partial differential equations (PDEs), e.g., families of PDEs with different source terms, boundary conditions, or initial conditions. Here, we adapt the approaches to address the particular use case of AAA by integrating the 3D Navier-Stokes equations (NSE) as the physical laws governing fluid dynamics. In addition, we follow best practices to enhance the capabilities of the models by effectively capturing the underlying physics of the problem under study. The advantages and limitations of each approach are highlighted through a series of relevant application cases. We validate our results by comparing them with CFD simulations for benchmark datasets, demonstrating good agreements and emphasizing those cases where improvements in computational efficiency are observed.  The proposed methodology serves as a starting point for future research in the application of Deep Learning in cardiovascular disease modeling and offers a promising alternative for real-time simulation and monitoring of vascular flow in clinical settings.

\end{abstract}


\end{frontmatter}

\newpage
\section{Introduction}\label{sect:intro}
Abdominal Aortic Aneurysms (AAA) are a critical health concern, characterized by the enlargement of the aorta diameter in the abdominal region. Accurate and efficient simulation of vascular flow within AAA is essential for predicting rupture risks and planning surgical interventions. 

Recent developments in magnetic resonance technology have led to the increased application of 4D flow MRI for analyzing hemodynamics. This imaging technique allows for the measurement of three-dimensional blood flow fields. Despite its benefits, the applicability to assess blood flow near vessel walls is challenged by limitations in spatial and temporal resolution, as well as the limited accuracy in analyzing slow-moving blood (see \cite{Jarral2020}).

On the other hand, computational methods such as Computational Fluid Dynamics (CFD), which rely on Navier-Stokes equations to simulate blood flow coupled with elastodynamic equations to model the arterial wall behavior have been widely adopted to model complex flow patterns in both large and small vessels with pathological changes. Using patient-specific inflow and outflow boundary conditions as well as patient-specific structure BCs, CFD simulations, taking into account fluid structure interaction (FSI), can deliver high-resolution insights into the elastohemodynamics analysis. Additionally, FSI enables the evaluation of pressure, shear stress and wall strains and stresses which are otherwise challenging to measure directly in living patients. Despite these benefits, the accuracy of these complex simulations is largely dependent on having access to detailed flow and structure BCs for all relevant vessels, which may be difficult to obtain for practical or ethical reasons. Moreover, CFD simulations accounting of such fluid structure interaction is time-intensive and require considerable computational resources, reducing their suitability for real-time clinical use \citep{Zhu2022}.

So then, it is crucial to explore new data assimilation algorithms that offer possible alternatives to these limitations. Deep learning to achieve super-resolution and denoising of 4D flow MRI data, prediction of CFD simulations or combining approaches to build improved models are emerging areas of research and development. For instance, the authors in \cite{Fathi2020} introduced a framework that embeds the governing Navier–Stokes equations into a deep neural network to simultaneously achieve super‐resolution and denoising without requiring high‐resolution labels. Building on this concept, \cite{Gao2021} developed a convolutional neural network approach that enforces physical constraints to recover high‐fidelity flow fields from low‐resolution inputs, further demonstrating the potential of label‐free training in practical imaging scenarios. \cite{Saitta2024} proposed an implicit neural representation method for unsupervised super‐resolution and denoising of 4D flow MRI data, offering a computationally efficient solution that inherently captures the underlying physics. Additionally in \cite{Liang2020}, machine learning techniques, without relying on physics-informed methods, were used to develop DNNs that directly estimate 3D steady-state pressure and flow velocity distributions in a simplified model of the thoracic aorta. Similarly, \cite{Li2021} applied a dual-channel deep learning network along with cardiovascular hemodynamic point datasets in 3D to model the relationship between cardiovascular geometry and internal hemodynamics. Statistical analysis confirmed that the deep learning predictions matched conventional CFD results, while reducing computational time. \cite{Ferdian2020} combined CFD simulations and deep learning to build a model that reduces noise and enhances the resolution of 4D Flow MRI velocities. Recent advances have also seen the results of \cite{Balzotti2022}, who introduced a data-driven approach to optimizing blood flow in coronary artery bypass grafts using a novel reduced-order model that combines proper orthogonal decomposition (POD) with neural networks. This method is applied to steady 3D flows, focusing on patient-specific geometries and addressing the challenge of varying Reynolds numbers within physiological ranges. In \cite{Garay2024}, the authors applied Physics-Informed Neural Networks (PINNs) to solve inverse hemodynamics problems in 3D, especially useful when boundary information is lacking and high-quality blood flow measurements are difficult to obtain. Focusing on the ascending aorta and taking advantage of sparse and noisy 2D measurements, the study demonstrated robust and relatively accurate parameter estimations when using the method with simulated data, while the velocity reconstruction accuracy shows dependence on the measurement quality and the flow pattern complexity. However, the study acknowledged its limitations, such as the assumption that the walls are rigid and the lack of validation with real clinical data. The reference solution is built using the finite element method (FEM). Additionally, in \cite{Zhu2024} further extensions to PINN were made to model incompressible flows in 2D for idealized geometries and time-dependent moving boundaries, a major challenge in current methodologies. By incorporating Dirichlet velocity constraints and refining the training points around the moving boundaries, the study improved the accuracy and applicability of no-slip conditions. This extended PINNs version demonstrated its effectiveness in solving unsteady flow problems and the potential for inverse problems, highlighting an area for future improvement. In \cite{Kashefi2022}, an innovative physics-informed deep learning framework was introduced with the aim to solve steady-state incompressible flows over 2D irregular geometries. By employing point cloud neural networks and formulating loss functions based on governing equations and sparse observations, the framework surpassed the limitations of traditional PINNs, allowing solutions across multiple computational domains with significant geometric variations. The adaptability of the methodology to unknown geometries promises significant computational savings and broad applicability in modeling irregular cardiovascular geometries. In \cite{Daneker2024}, the authors presented a new computational framework called WS-PINNs, where WS refers to Warm-Start, to improve hemodynamics analysis of type B aortic dissections in mouse models in 3D. In general, they explored the impact of spatial and temporal resolution, noisy data processing, and transfer learning to potentially improve prognostic capability and understand aneurysm development. In \cite{Yin2022}, the authors investigated the complex progression of aortic dissection through medial wall delamination employing DeepONet, an operator regression neural network, to develop a surrogate model that simulates the delamination process under various strut distributions. The study successfully predicted fluid injection pressure-volume curves and damage progression, providing insights into the mechanical impact of histological microstructures on dissection. Some of the limits imposed are the use of synthetic data and simplified 2D geometries.  In two recent works, the authors in \cite{Li2024A,Li2024B} proposed a framework based on the physics-informed DeepONet approach that employs a one-dimensional model of the Navier–Stokes equations for blood flow simulation. By incorporating time-periodic constraints, Windkessel boundary conditions, and meta-learning strategies for parameter estimation, these methods predict continuous arterial blood pressure waveforms in both space and time from readily measurable signals (e.g., ECG, PPG). Notably, only outlet boundary data are required, offering a promising route toward continuous BP monitoring in clinical scenarios.

As observed, a key development is provided by models such as Physics-Informed Neural Networks (PINNs) and Deep Operator Networks (DeepONets), which have proven effective in addressing various engineering problems, offering increased accuracy and efficiency in many scenarios. 

PINN is a method that combines deep learning with the fundamental principles of physics (see \cite{Raissi2017PartI, Raissi2017PartII, Raissi2019}). The main idea is to build deep neural networks, mainly using the multilayer perceptron (MLP) architecture, that incorporate into the loss function the physical laws governing the problem which are described by partial differential equations (PDEs) and include the governing equations, and the boundary and initial conditions. Unlike traditional deep learning methods, or end-to-end approaches, which rely on data to learn relationships between inputs and outputs, PINN takes advantage of this modified loss function to train the neural network. During training, the network adjusts its parameters not only to minimize errors in the training data but also to meet the constraints imposed by the physical equations. This approach makes PINN very effective for solving complex problems where data might be sparse or expensive to obtain, while also ensuring that the solutions respect the underlying physical laws (see \cite{Cai2021, Farea2024}). There is also a variant of PINN that works without any data, other than the boundary and initial conditions and the unsupervised constraints imposed by the governing equations, which is particularly useful for solving problems where experimental data or numerical simulations may not be available (see \cite{Jin2021, Sun2020}). It has been tested as an alternative ML solver to the classical CFD solvers, although most of them are currently in the development stages.
 
On the other hand, DeepONet is a technique designed to learn nonlinear operators from data (see \cite{LuLu2021DeepONet}). In contrast to traditional neural networks, which learn functions, i.e., mappings between inputs and outputs, DeepONet aims to learn operators, which are mappings between functions. This means DeepONet can learn to generalize complex transformations between input functions and output functions, which is particularly useful in studying parametric partial differential equations, e.g., families of PDEs with different source terms, boundary conditions, or initial conditions.  In addition, DeepONet is particularly powerful because it can make accurate predictions even with limited training data and generalize to unseen scenarios. While classical PINN requires retraining when the source terms or boundary conditions change (note that Parameterized-PINNs arise to address the issue \cite{Cho2024, Liu2024}), DeepONet handles this scenario without needing the repetitive and time-consuming retraining. In that way, the approach predicts new results under different conditions based on what it has previously learned, offering more flexibility and scalability compared to PINNs. There is also a generalization of DeepONet that allows incorporating the physical laws governing the problem as PINN does and is called PI-DeepONet (see \cite{Wang2021DeepOnets, Goswami2023}). This variant combines the advantages of both methods and is notable in scenarios without labeled data sets, where the residual PDE loss is used to learn the underlying physics. However, the approach encounters substantial computational challenges and is under active development (see \cite{Wang2022Improved, Mandl2024}). 

In this work, we apply Physics-Informed Neural Networks (PINNs), Deep Operator Networks (DeepONets), and their Physics-Informed extensions (PI-DeepONets) for predicting vascular flow simulations in the context of a 3D Abdominal Aortic Aneurysm (AAA) idealized model. This research aims to assess the impact of these methods on the accurate prediction of simulations and to evaluate potential improvements in computational efficiency over classical CFD approaches. Our benchmark study employs a simplified setup using a steady-state 3D Navier–Stokes model with a steady inflow profile. We acknowledge that this idealized model does not capture the full complexity of real AAA hemodynamics, such as patient-specific inflow conditions and pulsatile dynamics, and that advanced FSI simulations exist for detailed patient-specific modeling, though they are computationally expensive and rarely fully patient-specific. In this context, our approach aligns well with current PINN and PI-DeepONet methodologies, providing a practical and efficient baseline for further development.

The novelty of our work is highlighted by several key aspects that differentiate it from the existing literature:

\begin{enumerate}
\item Combined Study of PINN and (PI-)DeepONet\\
While previous works often focus on a single approach (e.g., PINNs or (PI-)DeepONets), our manuscript performs a comprehensive, side-by-side study of PINNs, DeepONets, and PI-DeepONets in the same AAA context. This provides a comparative benchmark for the relative strengths, limitations and applicability of each method. \\
In addition, we analyze these methods under diverse data conditions (e.g., sparse, noisy, fully data-driven, and purely physics-based) for a single 3D geometry, offering practical guidelines on selecting the most suitable approach in terms of accuracy, noise resilience, and computational cost for a given use case.

\item Use of a 3D Abdominal Aortic Aneurysm Model\\
To the best of the authors' knowledge, there are recent published studies applying PINNs to 3D cardiovascular flow (for instance, \cite{Fergus2023,Gu2024,Daneker2024,Garay2024}). However, some of the main differences with respect to our PINNs sections lie in the geometry of the problem under study, the specific techniques we employ to enhance the performance of PINNs, and the fact that we apply PINNs with and without loss data. It is worth mentioning that our Sections \ref{sect:results:pinns:drf} and \ref{sect:results:pinns:noise} are inspired by the ideas of \cite{Daneker2024}.\\
On the other hand, operator-based approaches have largely focused on one- or two-dimensional domains (e.g., \cite{Yin2022,Li2024A,Li2024B}). However, here we adapt DeepONets, including their physics-informed variants PI-DeepONets, to tackle a three-dimensional AAA idealized model, which is more complex than several prior simplified cases in the literature.

\item Multi-Input, Multi-Output (PI-)DeepONet Architecture\\
Based on the works of \cite{Jin2022MultipleInput} and \cite{Wang2023Long}, we propose a (PI-)DeepONet architecture that seamlessly integrates multiple inputs (e.g., varying maximum inlet velocity and outlet pressure) and simultaneously predicts multiple outputs (e.g., velocity and pressure fields) within a single operator-learning paradigm. This design is a step further from previous (PI-)DeepONet work to address large-scale 3D cardiovascular flows.

\item Implementation of Advanced Training Strategies\\
Our paper incorporates advanced training techniques that integrate best practices from recent literature to tackle common optimization challenges. This comprehensive approach ensures a reproducible benchmark that researchers and practitioners can follow when applying similar neural network models in fluid mechanics.

\item Inference Speed and Real-Time Feasibility\\
The results not only demonstrate strong agreement with high-fidelity CFD simulations but also reveal significant improvements in computational efficiency. In particular, the inference runtime in the (PI-)DeepONet approach achieves speedups of 22.5× with respect to CFD simulations, highlighting its potential for real-time applications.

\end{enumerate}

The manuscript is organized as follows. In Section 2, we present a background of PINNs and DeepONet for a general problem and introduce the fundamental notions of the paper. In Section 3, we formulate the Abdominal Aortic Aneurysm (AAA) idealized model. In Section 4, we adapt both PINNs and DeepONet to suit our specific use case, incorporating the Navier-Stokes equations (NSE) as the fundamental physical laws governing fluid dynamics. Finally, in Section 5, we present numerical results highlighting the advantages and limitations of each approach through several relevant scenarios. We validate our findings by comparing them with CFD simulations on benchmark datasets.
\section{Background}\label{sect:background}
The scope of this section is twofold. First, we present the PINNs method applied to a general partial differential equation. We define the loss function and discuss the training process. In the second part, we follow a similar approach for DeepONet, this time applied to parametric-PDEs. Here, we also discuss its generalization to PI-DeepONet.

\subsection{Physics-Informed Neural Networks (PINNs)}\label{sect:background:pinns}
Let us consider a general partial differential equation (PDE) defined on an open spatial domain $\mathscr{B} \subset \mathbb{R}^d$ with closure $\overline{\mathscr{B}}$ and boundary $\partial \mathscr{B}$. We also consider a temporal domain $[0, T]$. The PDE is given as follows,
\begin{subequations}
\begin{align}
	& \mathfrak{N}[u(\boldsymbol{x}, t)] = 0, \quad \boldsymbol{x} \in \mathscr{B}, \quad t \in (0, T], \label{eq:background:pinns:pde:a}\\
	& \mathfrak{B}[u(\boldsymbol{x}, t)] = 0, \quad \boldsymbol{x} \in \partial\mathscr{B}, \quad t \in (0, T], \label{eq:background:pinns:pde:b} \\
	& u(\boldsymbol{x}, 0) = u_0(\boldsymbol{x}), \quad \boldsymbol{x} \in \overline{\mathscr{B}}, \label{eq:background:pinns:pde:c}
\end{align}
\end{subequations}
where $u: \overline{\mathscr{B}} \times [0,T] \to \mathbb{R}$ is the solution function of the PDE, $\boldsymbol{x} \in \mathscr{B}$ denotes the spatial vector variable, and $t$ refers to time. In addition, $\mathfrak{N}[\cdot]$ and $\mathfrak{B}[\cdot]$ are non-linear spatial-temporal differential operators, and $u_0: \overline{\mathscr{B}} \to \mathbb{R}$ represents the initial condition function.

For the sake of the explanation, we assume that Eqs. \eqref{eq:background:pinns:pde:a}-\eqref{eq:background:pinns:pde:c} describe a well-posed problem. At this point, the goal of solving the PDE using Deep Neural Networks (DNN) could be addressed by means of an end-to-end learning approach, where inputs are independent spatial and temporal variables $\boldsymbol{x}, t$, and the target output is the solution function $u(\boldsymbol{x}, t)$, training the model on previously known solution data. However, in this way we overlook all the remaining information provided by Eqs.\eqref{eq:background:pinns:pde:a}-\eqref{eq:background:pinns:pde:c}, such as the governing equation, the boundary and initial condition. 

Based on this idea, the authors in \cite{Raissi2017PartI, Raissi2017PartII, Raissi2019} proposed a technique called Physics-Informed Neural Networks (PINNs). In particular, the solution function $u(\boldsymbol{x}, t)$ is approximated with a fully connected deep neural network (FCNN) that receives as inputs the spatial and temporal variables, and outputs the predicted value $\hat{u}_{\theta}(\boldsymbol{x}, t)$, where $\theta$ represents all trainable network parameters. However, in contrast to the previous end-to-end approach, the goal now is to optimize the parameters $\theta$ by minimizing a more comprehensive loss function, i.e. $\theta^* = \arg\min_{\theta} \mathcal{L}(\theta)$, where
\begin{align}
& \mathcal{L}(\theta) = \mathcal{L}_\text{phy}(\theta) + \mathcal{L}_\text{bc}(\theta) + \mathcal{L}_\text{ic}(\theta) + \mathcal{L}_\text{data}(\theta). \label{eq:background:pinns:loss} 
\end{align}

\begin{figure}[htbp]
\hspace{-1cm}
\includegraphics[scale=0.9]{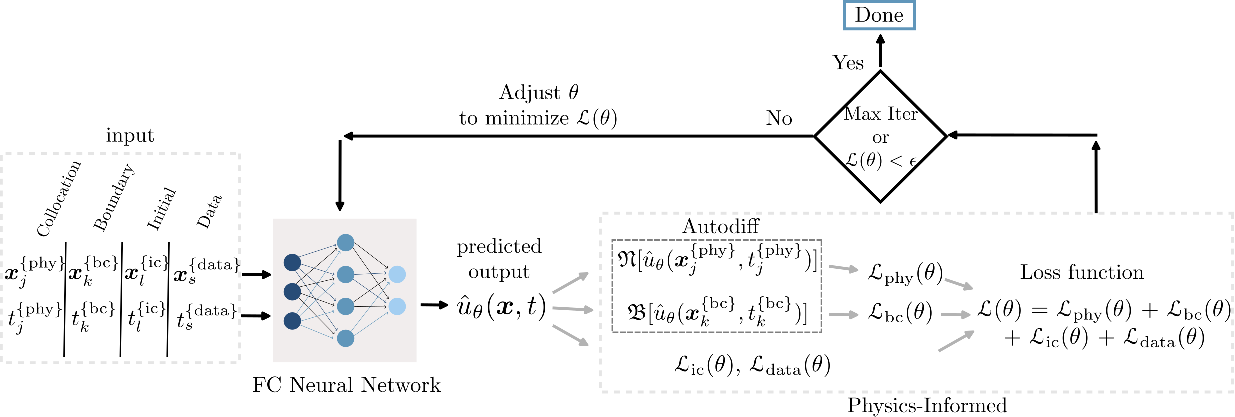}
\caption{Illustrative example of the PINNs training process.}
\label{fig:background:pinn}
\end{figure}

So then, the predicted output $\hat{u}_{\theta}(\boldsymbol{x}, t)$ must not only satisfy the previously known solution data, but also the PDE, and the boundary and initial conditions (see Figure \ref{fig:background:pinn}).  Notice that in Figure \ref{fig:background:pinn} the notation $\boldsymbol{x}^{\{\text{label}\}}_{j}$, $t^{\{\text{label}\}}_{j}$  refers to the $j$-th sample point in the region or stratum denoted by the label within the curly brackets \(\{\cdot\}\).

As observed, the training process focuses on the minimization of different loss terms of Eq. \ref{eq:background:pinns:loss} that enforce the underlying physics, conditions, and solution data of the problem. One key loss term is the Physics Loss, denoted by $\mathcal{L}_\text{phy}$, which ensures that the residual of the PDE is minimized at collocation points $ \{ (\boldsymbol{x}^{\{\text{phy}\}}_{j}, t^{\{\text{phy}\}}_{j}) \}_{j=1}^{P_\text{phy}}$ in the domain (see Eq. \eqref{eq:background:pinns:pde:a}). It is defined as,
\begin{align}
& \mathcal{L}_\text{phy}(\theta) = \frac{1}{P_\text{phy}} \sum_{j=1}^{P_\text{phy}} \left| \mathfrak{N}[\hat{u}_{\theta}(\boldsymbol{x}^{\{\text{phy}\}}_{j}, t^{\{\text{phy}\}}_{j})]  \right|^2. \label{eq:background:pinns:lossphy}
\end{align}

Additionally, the Boundary Condition Loss that refers as $\mathcal{L}_\text{bc}$, enforces the boundary conditions at points $ \{(\boldsymbol{x}^{\{\text{bc}\}}_{k}, t^{\{\text{bc}\}}_{k}\}_{k=1}^{P_\text{bc}}$ on the boundary (see Eq. \eqref{eq:background:pinns:pde:b}), and is expressed as follows,
\begin{align}
&\mathcal{L}_\text{bc}(\theta) = \frac{1}{P_\text{bc}} \sum_{k=1}^{P_\text{bc}} \left| \mathfrak{B}[\hat{u}_{\theta}(\boldsymbol{x}^{\{\text{bc}\}}_{k}, t^{\{\text{bc}\}}_{k})] \right|^2. \label{eq:background:pinns:lossbc}
\end{align}

The Initial Condition Loss $\mathcal{L}_\text{ic} $ ensures that the predicted solution satisfies the initial conditions at $ t = 0 $ (see Eq. \eqref{eq:background:pinns:pde:c}), formulated as,
\begin{align}
& \mathcal{L}_\text{ic}(\theta) = \frac{1}{P_\text{ic}} \sum_{l=1}^{P_\text{ic}} \left| \hat{u}_{\theta}(\boldsymbol{x}^{\{\text{ic}\}}_{l}, 0) - u_0(\boldsymbol{x}^{\{\text{ic}\}}_{l}) \right|^2, \label{eq:background:pinns:lossic}
\end{align}
where $ \{ \boldsymbol{x}^{\{\text{ic}\}}_{l}, t^{\{\text{ic}\}}_{l}=0 \}_{l=1}^{P_\text{ic}} $ are points in the domain at the initial time. 

Finally, the Data Loss denoted by $\mathcal{L}_\text{data}$, ensures that the predicted solution fulfills the known solution data at points $ \{ \boldsymbol{x}^{\{\text{data}\}}_{s}, t^{\{\text{data}\}}_{s} \}_{s=1}^{P_\text{data}} $, and is given as follows,
\begin{align}
	& \mathcal{L}_\text{data}(\theta) = \frac{1}{P_\text{data}} \sum_{s=1}^{P_\text{data}} \left| \hat{u}_{\theta}(\boldsymbol{x}^{\{\text{data}\}}_{s}, t^{\{\text{data}\}}_{s}) - u(\boldsymbol{x}^{\{\text{data}\}}_{s}, t^{\{\text{data}\}}_{s}) \right|^2. \label{eq:background:pinns:lossdata}
\end{align}

Notice that automatic differentiation (see \cite{Baydin2017, Paszke2017}) is used to compute the necessary derivatives of the neural network prediction $ \hat{u}_{\theta}(\boldsymbol{x}, t)$ with respect to spatial and temporal variables in $\mathfrak{N}[\cdot]$ and $\mathfrak{B}[\cdot]$ for losses $\mathcal{L}_\text{phy}(\theta)$ and $\mathcal{L}_\text{bc}(\theta)$, respectively (see Autodiff highlighted in Figure \ref{fig:background:pinn}).

\subsection{Operator Learning}\label{sect:background:operatorlearning}
Some parameters of a given PDE system are allowed to change in a given range, e.g., source term, domain shape, initial and boundary conditions, coefficients, etc. Classical PINNs require retraining when this occurs, which is disadvantageous and computationally expensive. A possible alternative is provided by Operator Learning. This section presents an overview of the Operator Learning paradigm, in particular, we discuss the DeepONet method and its extension to PI-DeepONet.

Let us consider the following Parametric-PDEs as a generalization to Eqs. \eqref{eq:background:pinns:pde:a}-\eqref{eq:background:pinns:pde:c} of the previous section,
\begin{subequations}
\begin{align}
	&\mathfrak{N}[u^{(i)}(\boldsymbol{x}, t); f^{(i)}(\boldsymbol{x})] = 0, \quad \boldsymbol{x} \in \mathscr{B}, \quad t \in (0, T], \label{eq:background:operator:pde:a}\\
	&\mathfrak{B}[u^{(i)}(\boldsymbol{x}, t)] = 0, \quad \boldsymbol{x} \in \partial\mathscr{B}, \quad t \in (0, T], \label{eq:background:operator:pde:b}\\
	&u^{(i)}(\boldsymbol{x}, 0) = u_0(\boldsymbol{x}), \quad \boldsymbol{x} \in \overline{\mathscr{B}}, \label{eq:background:operator:pde:c}
\end{align}
\end{subequations}
where $ \{ f^{(i)}(\boldsymbol{x}) \}_{i=1}^{N}$ is a family of $N$ different source terms in the form $f^{(i)}: \mathscr{B} \to \mathbb{R}$, and $u^{(i)}(\boldsymbol{x}, t)$ is the associated PDE solution function. For the explanation, we will use the source term function as the input parameter.

Operator Learning involves learning nonlinear operators that map from one infinite-dimensional Banach space to another using data. This paradigm is gaining significant attention in the context of solving PDEs (see \cite{Lanthaler2022}). Let's consider two separated Banach spaces, $\mathscr{X}, \mathscr{Y}$. Suppose we have a collection of input functions $\{ f^{(i)}(\boldsymbol{x})\}_{i=1}^{N}$, where each $f^{(i)} \in \mathscr{X}$, and their corresponding output functions $\{ u^{(i)}(\boldsymbol{x}, t)\}_{i=1}^{N}$, where each $u^{(i)} \in \mathscr{Y}$.

The aim is to learn an operator $\mathfrak{G}$: $\mathscr{X} \mapsto \mathscr{Y}$ such that for any input function $f^{(i)}$, we can compute the corresponding output function $u^{(i)} = \mathfrak{G}(f^{(i)})$. In the context of Eqs. \eqref{eq:background:operator:pde:a}-\eqref{eq:background:operator:pde:c}, the operator $\mathfrak{G}$ outputs the solution function $u^{(i)}$ to the PDE for a given input function $f^{(i)}$. Additionally, it follows that for all points $(\boldsymbol{x}, t)$ in the domain of the solution, the following relationship holds true,
\begin{align}
	& u^{(i)}(\boldsymbol{x}, t)=\mathfrak{G}(f^{(i)})(\boldsymbol{x}, t). \label{eq:background:operatorlearning}
\end{align}

For this purpose, Neural Networks are proposed to approximate the operator $\mathfrak{G}$. Let's dive into DeepONet. 
\subsubsection{Deep Operator Network (DeepONet)}\label{sect:background:deeponet}

\begin{figure}[htbp]
	\centering	
	\includegraphics[scale=0.8]{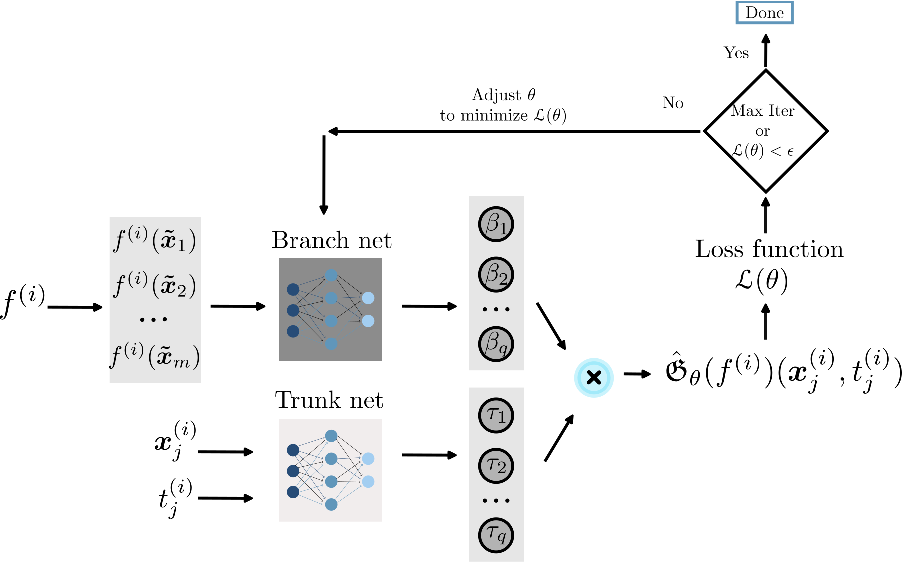}
	\caption{DeepONet workflow.}
	\label{fig:background:deeponet}
\end{figure}

The authors in \cite{LuLu2021DeepONet} proposed to represent the solution map $\mathfrak{G}$ by an unstacked deep learning architecture called DeepONet (see Figure \ref{fig:background:deeponet}). As observed, DeepONet is composed of two neural networks referred to as the Branch net and the Trunk net, respectively. The goal is to approximate $\mathfrak{G}$ by means of the DeepONet output $\hat{\mathfrak{G}}_{\theta}$, defined as follows,
\begin{align}
& \hat{\mathfrak{G}}_\theta(f^{(i)})(\boldsymbol{x}^{(i)}_j, t^{(i)}_j):=\sum_{k=1}^q\underset{Branch}{\underbrace{{\beta}_{k}\left(f^{(i)}(\tilde{\boldsymbol{x}}_1),f^{(i)}(\tilde{\boldsymbol{x}}_2),...,f^{(i)}(\tilde{\boldsymbol{x}}_m)\right)}}\;\underset{Trunk}{\underbrace{\tau_k(\boldsymbol{x}^{(i)}_j, t^{(i)}_j)}}, \label{eq:background:deeponet:definition}
\end{align}
where $\theta$ denotes the set of all trainable parameters in both branch and trunk networks. DeepONet architecture is grounded in the Universal Approximation Theorem for Operators, which states that neural networks of this form can approximate continuous nonlinear operators to arbitrary accuracy under certain conditions (see \cite{Chen1995, LuLu2021DeepONet}).

Specifically, the Branch network encodes input functions into a compact representation or feature embedding $[\beta_1, \beta_2, ..., \beta_q]^{T}\in \mathbb{R}^q$, where $[f^{(i)}(\tilde{\boldsymbol{x}}_1),f^{(i)}(\tilde{\boldsymbol{x}}_2),...,f^{(i)}(\tilde{\boldsymbol{x}}_m)]^{T}$ represents the function $f^{(i)}$ evaluated at a collection of $m$ sensor points $ \{ \tilde{\boldsymbol{x}}_{k} \}_{k=1}^{m} \subset \mathscr{B}$. It is worth pointing out that this collection is fixed, i.e., it remains the same across different input samples $f^{(i)}$. On the other hand, the Trunk network maps the spatial and temporal coordinate points into a feature embedding $[\tau_1, \tau_2, ..., \tau_q]^{T}\in \mathbb{R}^q$. To obtain the final predicted output $\hat{\mathfrak{G}}_\theta(f^{(i)})(\boldsymbol{x}^{(i)}_j, t^{(i)}_j)\in \mathbb{R}$, the networks are merged via a dot product (see Eq. \eqref{eq:background:deeponet:definition}).

DeepONet is a purely data-driven model that during training adjusts all trainable parameters $\theta$ to minimize the error between model predictions and actual data. This can be enforced into a loss function as follows,
\begin{align}
	& \mathcal{L}(\theta)=\frac{1}{NP}\sum_{i=1}^N\sum_{j=1}^P\left|\hat{\mathfrak{G}}_{\theta}(f^{(i)})(\boldsymbol{x}_j^{(i)}, t_j^{(i)})-\mathfrak{G}(f^{(i)})(\boldsymbol{x}_j^{(i)}, t_j^{(i)})\right|^2, \label{eq:background:deeponet:generalloss}
\end{align}
where $N$ is the number of input functions in our training dataset, and $P$ is the number of points in the domain of the solution function. 

So then, by combining Eq. \eqref{eq:background:deeponet:definition} and Eq.\eqref{eq:background:deeponet:generalloss}, the loss function is rewritten as follows,
\begin{align}
	& \mathcal{L}(\theta)=\frac{1}{NP}\sum_{i=1}^N\sum_{j=1}^P\left|\sum_{k=1}^q {\beta}_{k}\left(f^{(i)}(\tilde{\boldsymbol{x}}_1),f^{(i)}(\tilde{\boldsymbol{x}}_2),...,f^{(i)}(\tilde{\boldsymbol{x}}_m)\right)\cdot \tau_k(\boldsymbol{x}^{(i)}_j, t^{(i)}_j)-\mathfrak{G}(f^{(i)})(\boldsymbol{x}^{(i)}_j, t^{(i)}_j)\right|^2. \label{eq:background:deeponet:loss}
\end{align}

Notice that the dataset for training a DeepONet architecture is given as a triplet with inputs $(\boldsymbol{x},t)$ and $f$, and output $\mathfrak{G}(f)(\boldsymbol{x}, t)$. Their respective dimensions are $(N\times P, d+1)$, $(N\times P, m)$ and $(N\times P, 1)$ (see \cite{Wang2021DeepOnets}). Please, refer to \ref{appendix:deeponet:dataset:dim} for further exploration.

\subsubsection{Physics-Informed Deep Operator Network (PI-DeepONet)}\label{sect:background:pideeponet}
The authors in \cite{Wang2021DeepOnets} combined the comprehensive loss function in PINNs and the network architecture of DeepONet, and proposed a generalized method called Physics-Informed Deep Operator Networks (PI-DeepONet). This approach incorporates physical laws into DeepONet to ensure that the learned operator $\hat{\mathfrak{G}}_\theta$ not only fits the data but also satisfies the governing equations, and the initial and boundary conditions of the PDEs.

The workflow in PI-DeepONet consists the optimization of the parameter $\theta$ by minimizing a total loss $\mathcal{L}(\theta)$, similar to the one given in Eq. \eqref{eq:background:pinns:loss}, 
\begin{align}
	& \mathcal{L}(\theta) = \mathcal{L}_\text{phy}(\theta) + \mathcal{L}_\text{bc}(\theta) + \mathcal{L}_\text{ic}(\theta) + \mathcal{L}_\text{data}(\theta). \label{eq:background:pideeponet:loss} 
\end{align}

However, in contrast to Eqs. \eqref{eq:background:pinns:lossphy}-\eqref{eq:background:pinns:lossdata}, the loss functions are now adapted to take into account the predicted output $\hat{\mathfrak{G}}_\theta$. Thus, we have the following expressions,
\begin{subequations}
\begin{align}
	& \mathcal{L}_\text{phy}(\theta)=\frac{1}{NP_\text{phy}}\sum_{i=1}^N\sum_{j=1}^{P_\text{phy}}\left| \mathfrak{N}\left[\hat{\mathfrak{G}}_\theta(f^{(i)})(\boldsymbol{x}^{\{\text{phy}\}(i)}_{j}, t^{\{\text{phy}\}(i)}_{j});f^{(i)}(\boldsymbol{x}^{\{\text{phy}\}(i)}_{j})\right]\right|^2, \label{eq:background:pideeponet:lossphy}\\
	&  \mathcal{L}_\text{bc}(\theta)=\frac{1}{NP_\text{bc}}\sum_{i=1}^N\sum_{k=1}^{P_\text{bc}}\left|\mathfrak{B}\left[\hat{\mathfrak{G}}_{\theta}(f^{(i)})(\boldsymbol{x}_{k}^{\{\text{bc}\}(i)}, t_{k}^{\{\text{bc}\}(i)})\right]\right|^2, \label{eq:background:pideeponet:lossbc}\\
	& \mathcal{L}_\text{ic}(\theta)=\frac{1}{NP_\text{ic}}\sum_{i=1}^N\sum_{l=1}^{P_\text{ic}}\left|\hat{\mathfrak{G}}_{\theta}(f^{(i)})(\boldsymbol{x}_{l}^{\{\text{ic}\}(i)}, 0)-u_0(\boldsymbol{x}_{l}^{\{\text{ic}\}(i)})\right|^2, \label{eq:background:pideeponet:lossic}\\
	& \mathcal{L}_\text{data}(\theta)=\frac{1}{NP_\text{data}}\sum_{i=1}^N\sum_{s=1}^{P_\text{data}}\left|\hat{\mathfrak{G}}_{\theta}(f^{(i)})(\boldsymbol{x}_{s}^{\{\text{data}\}(i)}, t_{s}^{\{\text{data}\}(i)})-u^{(i)}(\boldsymbol{x}_{s}^{\{\text{data}\}(i)}, t_{s}^{\{\text{data}\}(i)})\right|^2, \label{eq:background:pideeponet:lossdata}
\end{align}
\end{subequations}
where $ \{ f^{(i)} \}_{i=1}^{N} $ denotes a set of input functions. For each $f^{(i)}$, sets $ \{ (\boldsymbol{x}_{j}^{\{\text{phy}\}(i)}, t_{j}^{\{\text{phy}\}(i)}) \}_{j=1}^{P_\text{phy}} $, $ \{ (\boldsymbol{x}_{k}^{\{\text{bc}\}(i)}, t_{k}^{\{\text{bc}\}(i)}) \}_{k=1}^{P_\text{bc}} $, $ \{ (\boldsymbol{x}_{l}^{\{\text{ic}\}(i)}, t_{l}^{\{\text{ic}\}(i)} = 0) \}_{l=1}^{P_\text{ic}} $, and $ \{ (\boldsymbol{x}_{s}^{\{\text{data}\}(i)}, t_{s}^{\{\text{data}\}(i)}) \}_{s=1}^{P_\text{data}} $ represent the respective collocation, boundary, initial condition, and data points, sampled from their corresponding region or stratum in the spatio-temporal domain. Note that these sets of points may vary across different input samples $f^{(i)}$.

\section{Formulation of the Abdominal Aortic Aneurysm Idealized Model}\label{sect:formulation}
\begin{figure}[htbp]
	\centering	
	\resizebox{\columnwidth}{!}{\includegraphics[scale=0.5]{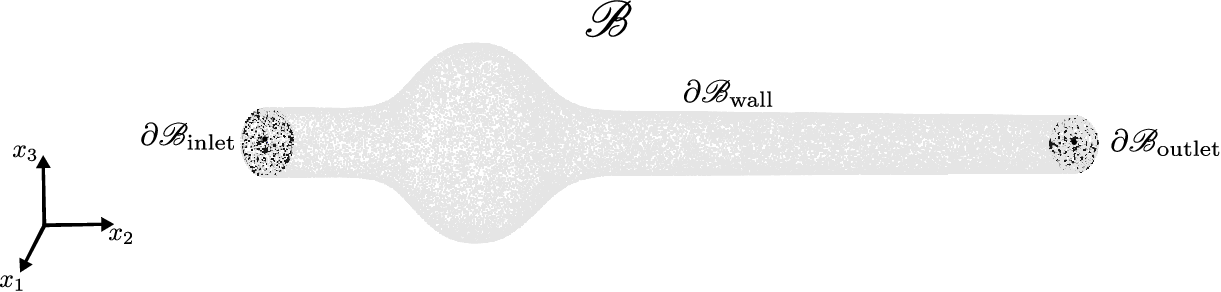}}
	\caption{Abdominal Aortic Aneurysm geometry. }
	\label{fig:formulation:aaa}
\end{figure}

In this work, we study the blood flow in 3D Abdominal Aortic Aneurysm (AAA) idealized model (see Fig \ref{fig:formulation:aaa}). The steady and incompressible flow was assumed to be laminar and the fluid behaved as Newtonian. The equations to solve were the Navier-Stokes ones. They are defined on a domain $\mathscr{B} \subset \mathbb{R}^3$  and are given in vectorial form as follows,
\begin{subequations}
	\begin{align}
		& \rho_f\left(\boldsymbol{v}(\boldsymbol{x}) \cdot \nabla\right) \boldsymbol{v}(\boldsymbol{x} ) = - \nabla p(\boldsymbol{x} ) + \mu_f \nabla^2\boldsymbol{v}(\boldsymbol{x} ), &&  \boldsymbol{x} \in \mathscr{B}, \label{eq:formulation:nse:momentum3D}\\
		& \nabla\cdot\boldsymbol{v}(\boldsymbol{x}) = 0, &&  \boldsymbol{x} \in \mathscr{B}. \label{eq:formulation:nse:mass3D}\\
		& \text{Boundary conditions} \nonumber\\
		& \boldsymbol{v}(\boldsymbol{x}) = \boldsymbol{v}^{\{\text{inlet}\}}(\boldsymbol{x}), &&  \boldsymbol{x} \in \partial\mathscr{B}_{\text{inlet}}, \label{eq:formulation:nse:bc:inlet}\\
		& \boldsymbol{v}(\boldsymbol{x}) = \boldsymbol{0}, &&  \boldsymbol{x} \in \partial\mathscr{B}_{\text{wall}}, \label{eq:formulation:nse:bc:wall}\\
		& \frac{\partial\boldsymbol{v}(\boldsymbol{x})}{\partial \boldsymbol{n}} = \boldsymbol{0}, &&  \boldsymbol{x} \in \partial\mathscr{B}_{\text{outlet}}. \label{eq:formulation:nse:bc:outlet}
	\end{align}
\end{subequations}

Eq. \eqref{eq:formulation:nse:momentum3D} corresponds to the conservation of momentum and Eq. \eqref{eq:formulation:nse:mass3D} represents the conservation of mass. There are three independent variables which are the spatial coordinates $x_1$, $x_2$, and $x_3$, and four dependent variables, the pressure scalar field $p := p(x_1,x_2,x_3)$ and the three components of the velocity vector field $\boldsymbol{v} := (v_1(x_1,x_2,x_3),v_2(x_1,x_2,x_3),v_3(x_1,x_2,x_3))$. Additionally, $\rho_f$ refers to the density and $\mu_f$ is the dynamic viscosity of the fluid both were constants.

We have also prescribed boundary conditions on the main limit surfaces, i.e., inlet, wall and outlet, and hence the solution of velocity and pressure can be uniquely determined. Specifically, a parabolic profile for the velocity is imposed for the boundary conditions at the inlet (see Eq. \eqref{eq:formulation:nse:bc:inlet}), with the inflow directed perpendicular to the inlet region. Thus, the inlet velocity at any coordinate point $\boldsymbol{x} \in \partial\mathscr{B}_{\text{inlet}}$ is given by $\boldsymbol{v}^{\{\text{inlet}\}}(\boldsymbol{x}):=(0,v^{\{\text{inlet}\}}_{2}(\boldsymbol{x}),0)$ where
\begin{align}
	v^{\{\text{inlet}\}}_{2}(\boldsymbol{x}) = V\left(1 - \frac{r^2(\boldsymbol{x})}{R^2}\right), \label{eq:formulation:inlet:parabolic}
\end{align}
and $r(\boldsymbol{x})$ represents the distance from the coordinate point to the inlet center, $R$ is the inlet radius and $V$ is the maximum inlet velocity. As observed for Poiseuille velocity profile, the velocity of the inlet region is maximum at the center and decreases to zero toward the edge. On the other hand, a non-slip condition is assumed on the geometry wall (see Eq. \eqref{eq:formulation:nse:bc:outlet}), which means that velocity are all set to zero, and the outflow boundary condition is imposed in the outlet (see Eq. \eqref{eq:formulation:nse:bc:outlet}).

Figure \ref{fig:formulation:aaa} illustrates the idealized geometry of the AAA model under study. The specimen is designed in such a way that the center of the inlet region corresponds to the origin of the Cartesian coordinate system and the main flow direction is through the $x_2$-axis. The fluid properties and geometrical parameters that are essential for solving the Navier-Stokes equations (NSE) are shown in Table \ref{tab:formulation:parameters}. Note that we have a set of eight different maximum inlet velocities ($V$).

 \begin{table}[htbp]
	 \centering
	\resizebox{16.5cm}{!}{%
		\begin{tabular}{c|c|c|c|c}
			     \hline
			     $\mu_f$ $[kg/(m s)]$& $\rho_f$ $[kg/m^3]$  &  $V$ $[m/s]$ &  $R$ $[m]$ & specimen length $[m]$ \\\hline
			     0.00399& 1060& [0.04, 0.05, 0.06, 0.08, 0.1, 0.12, 0.13, 0.15] & 0.010065 &0.26009 \\\hline
			 \end{tabular}}
	 \caption{NSE parameters. }
	 \label{tab:formulation:parameters}
\end{table}

\subsection{Ground Trust CFD simulations}\label{sect:cfd}

Fluent fluid solver (finite volume method) from Ansys Workbench R2 2020 (ANSYS Inc, Canonsburg, USA), was used to perform all the CFD simulations. The fluid domain was discretized in 1186076 elements with a layer of 7 prismatic elements in the near wall to capture the velocity gradients in the boundary layer. Concerning the numerical schemes, the pressure–velocity linkage was resolved by adopting a coupled algorithm that solves the momentum and pressure-based continuity equations together, the Least-Squares Cell-Based (LSCB) algorithm was used for spatial discretization of gradients. The pressure was solved through a second order scheme for momentum. Convergence was assumed when the root mean square residual error was less than 0.0001, within a maximum of 100 iterations. This approach takes about $3$ minutes per $V$-dependent CFD simulation and the data storage is about 118 MB.
\section{Adapting PINNs and DeepONet for AAA Simulations}\label{sect:adapting}

This section covers the data source, preprocessing steps, and the generation of training, validation and test datasets. In addition, here, we discuss how to adapt the classical PINNs and (PI-)DeepONet methods described in Section \ref{sect:background} to be applied to predicting velocities and pressure of the steady fluid in the AAA idealized model.

\subsection{Dataset Preparation}\label{sect:adapting:dataset}
\begin{figure}[htbp]
	\centering	
	\includegraphics[scale=0.6]{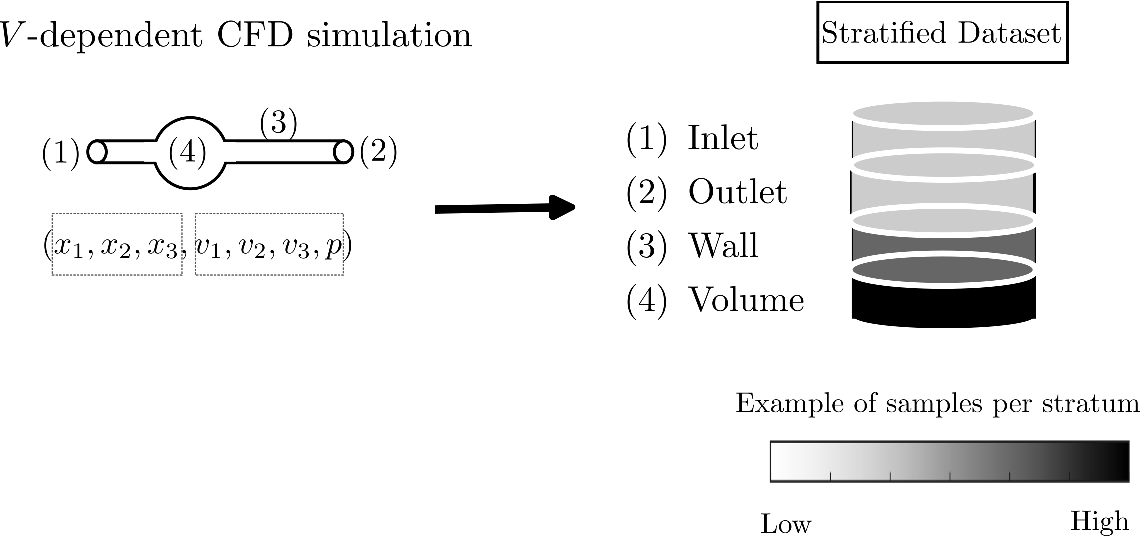}
	\caption{Schematic representation of a $V$-dependent stratified dataset.}
	\label{fig:adapting:datapreparation}
\end{figure}

The ground trust dataset consists of the solution of Navier-Stokes equations \eqref{eq:formulation:nse:momentum3D}-\eqref{eq:formulation:nse:bc:outlet} for the set of eight different maximum inlet velocities ($V$) displayed in Table \ref{tab:formulation:parameters} via CFD simulations (see Section \ref{sect:cfd}). In particular, we retrieve the 3D mesh coordinate points with their associated velocities and pressures. Then, we consider four main strata that are related to the geometry of the problem and will be used in the conception of the total loss function for PINNs and (PI-)DeepONet, respectively.  These regions include the Inlet, Outlet, Wall, and the Volume of the domain itself.  Thus, each $V$-dependent dataset within the ground trust dataset is assumed to be stratified, ensuring that access to $(x_1,x_2,x_3,v_1,v_2,v_3,p)$ is available per stratum (see Figure \ref{fig:adapting:datapreparation}).

In the literature, we can find use cases in which a closed-form solution to the studied problem or a simple-to-apply numerical method is available. In those cases, the performance of PINN and (PI-)DeepONet models can be respectively assessed by comparing their predictions at new points generated from the problem domain (in-domain resampling) or by randomly sample different input functions from a particular function space to finally generate the corresponding test dataset.  However, in our case, the ground trust dataset is immutable, and therefore, no resampling is considered.

\begin{figure}[htbp]
	\centering	
	\includegraphics[scale=0.45]{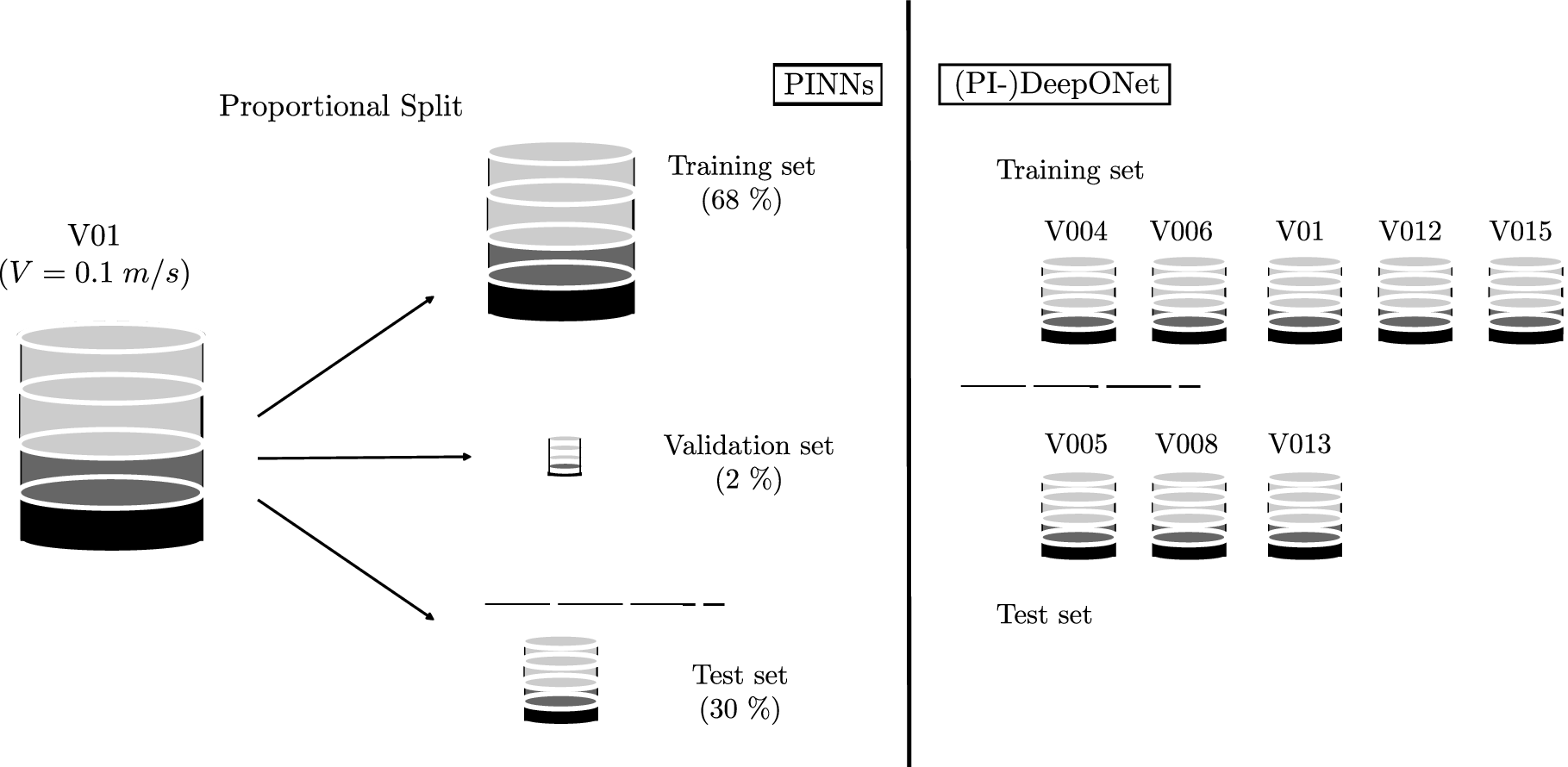}
	\caption{Schematic representation of the dataset splitting process.}
	\label{fig:adapting:split}
\end{figure}

So then, a good practice for PINNs here is to split each $V$-dependent stratified dataset into training/validation/test sets. In particular, we perform a stratified split, in which the percentage of data devoted to training, validation and testing is set to $68\%$, $2\%$ and $30\%$, respectively. In the case of (PI-)DeepONet, we split the ground truth dataset into two groups, i.e., one for training the model and the other for testing (see Figure \ref{fig:adapting:split}).

It is worth mentioning that in the training phase of both models, we purposely considered the use of only a subset of the data from the different regions of the domain, although we have access to all velocity and pressure data at each point of the mesh. In the following,  this selective data usage is denoted with the aster ($^{*}$) superscript, that is (Inlet$^{*}$, Outlet$^{*}$, Wall$^{*}$, Volume$^{*}$). Specifically, we assume the following scenario for the training set: at the Inlet$^{*}$, the spatial coordinates \((x_1, x_2, x_3)\) and velocity components \((v_1, v_2, v_3)\) are provided, but the pressure value is unavailable. This is represented as \((x_1, x_2, x_3, v_1, v_2, v_3, \text{NaN})\), where NaN denotes missing data. A similar scenario applies to the Outlet$^{*}$. At the Wall$^{*}$, the data is represented as \((x_1, x_2, x_3, 0, 0, 0, \text{NaN})\). These three regions are involved in the boundary conditions (see Eqs. \eqref{eq:formulation:nse:bc:inlet}-\eqref{eq:formulation:nse:bc:outlet}). For points in the Volume$^{*}$, neither the velocity nor the pressure values are provided, which is represented as \((x_1, x_2, x_3, \text{NaN}, \text{NaN}, \text{NaN}, \text{NaN})\). Collocation points are selected as random subsets from the combined set of coordinate points within Input$^{*}$, Wall$^{*}$, Output$^{*}$, and Volume$^{*}$ regions.

\begin{figure}[htbp]
	\centering	
	\includegraphics[scale=0.8]{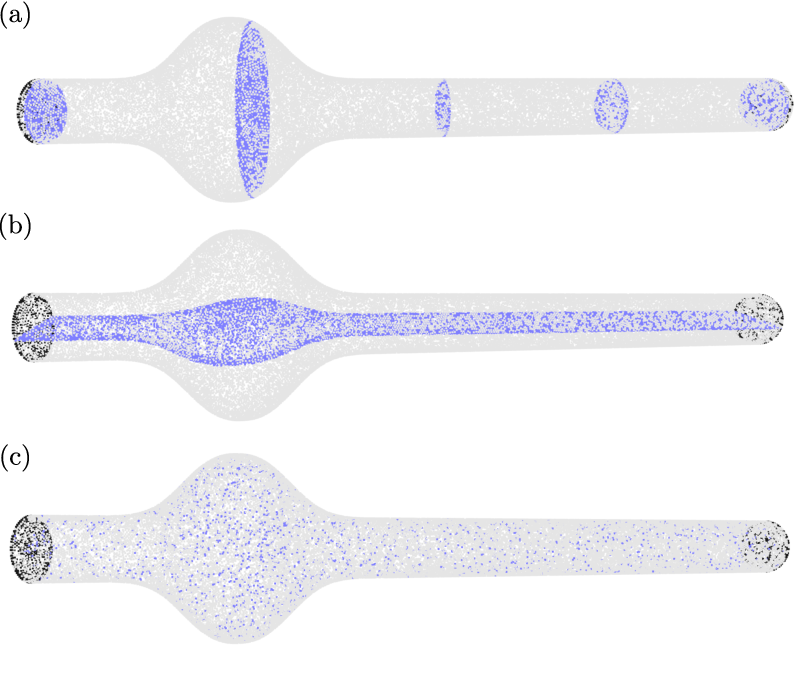}
	\caption{Mimicking the external Data set for three scenarios: (a) cross-section, (b) longitudinal and (c) random sample.}
	\label{fig:adapting:datascenarios}
\end{figure}

Additionally, we consider a subset of coordinate points within the Volume region in the training set where both velocity and pressure values are known. These points are conceived to mimic an external source of known Data that is used to train the model. This subset is constructed in three different ways, as illustrated in Figure \ref{fig:adapting:datascenarios}, by examining `cross-sectional', `longitudinal', and `random' sample cases. 

This selective data usage during training is designed to reflect real-world scenarios, where data in the physical domain is often sparse or unavailable. In such cases, models such as PINNs and (PI-)DeepONet are expected to infer the missing information based on the underlying physical laws governing the problem.

\subsection{PINNs Architecture and Loss Functions}\label{sect:adapting:pinns}

We use a PINN architecture consisting of a fully connected neural network that takes the coordinate points $(x_1, x_2, x_3)$ as inputs and simultaneously outputs the predicted values for the three components of the velocity field and the pressure field $(\hat{v}_{\theta 1}, \hat{v}_{\theta 2}, \hat{v}_{\theta 3}, \hat{p}_{\theta})$.

In addition, based on the ideas discussed in Sections \ref{sect:formulation} and \ref{sect:adapting:dataset}, we consider a total loss function ($\mathcal{L}$) that relies on the Navier-Stokes equations (see Eqs. \eqref{eq:formulation:nse:momentum3D}-\eqref{eq:formulation:nse:mass3D}), boundary conditions on Inlet, Wall and Outlet regions (see Eqs. \eqref{eq:formulation:nse:bc:inlet}-\eqref{eq:formulation:nse:bc:outlet}), and known data points within Volume (mimicking the external Data set). In this paper, we aim to show the benefits of PINNs by incorporating differential equations into the learning process to predict PDE solutions, especially when data are sparse or noisy, in contrast to methods that only learn from data. For this purpose, we make a distinction between a total loss function for a standard DeepNN, which utilizes only the known solution data through $\mathcal{L}_{\text{data}}$ and incorporates boundary conditions via $\mathcal{L}_{\text{inlet}}, \mathcal{L}_{\text{outlet}}, \mathcal{L}_{\text{wall}}$, and a total loss function for PINNs, which additionally incorporates the Navier-Stokes equations through $\mathcal{L}_{\text{phy}}$. That is,
\begin{subequations}
\begin{align}
    & \mathcal{L}_{\text{DeepNN}}(\theta) = \mathcal{L}_{\text{data}}(\theta) +  \mathcal{L}_{\text{inlet}}(\theta) +  \mathcal{L}_{\text{outlet}}(\theta) + \mathcal{L}_{\text{wall}}(\theta), \label{eq:adapting:deepnn:loss}\\
    & \mathcal{L}_{\text{PINN}}(\theta) = \mathcal{L}_{\text{data}}(\theta) +  \mathcal{L}_{\text{inlet}}(\theta) + \mathcal{L}_{\text{outlet}}(\theta) +  \mathcal{L}_{\text{wall}}(\theta) + \mathcal{L}_{\text{phy}}(\theta). \label{eq:adapting:pinns:loss}
\end{align}
\end{subequations}

Specifically, the loss components can be defined as, 
\begin{subequations}
\begin{align}
& \mathcal{L}_{\text{data}}(\theta) = \frac{1}{P_{\text{data}}}\sum_{j=1}^{P_{\text{data}}} \left( \left\| \hat{\boldsymbol{v}}_{\theta}(\boldsymbol{x}^{\{\text{data}\}}_{j}) - \boldsymbol{v}(\boldsymbol{x}^{\{\text{data}\}}_{j})\right\|^2 + \left| \hat{p}_{\theta}(\boldsymbol{x}^{\{\text{data}\}}_{j}) - p(\boldsymbol{x}^{\{\text{data}\}}_{j})\right|^2 \right),\label{eq:adapting:pinns:lossdata}\\
&\mathcal{L}_{\text{inlet}}(\theta) = \frac{1}{P_{\text{inlet}}} \sum_{k=1}^{P_{\text{inlet}}} \left\| \hat{\boldsymbol{v}}_{\theta}(\boldsymbol{x}^{\{\text{inlet}\}}_{k}) - \boldsymbol{v}(\boldsymbol{x}^{\{\text{inlet}\}}_{k})\right\|^2,\label{eq:adapting:pinns:lossinlet}\\ 
&\mathcal{L}_{\text{wall}}(\theta) = \frac{1}{P_{\text{wall}}}\sum_{l=1}^{P_{\text{wall}}} \left\| \hat{\boldsymbol{v}}_{\theta}(\boldsymbol{x}^{\{\text{wall}\}}_{l}) - \boldsymbol{v}(\boldsymbol{x}^{\{\text{wall}\}}_{l})\right\|^2, \label{eq:adapting:pinns:losswall}\\  
&\mathcal{L}_{\text{outlet}}(\theta) = \frac{1}{P_{\text{outlet}}}\sum_{s=1}^{P_{\text{outlet}}} \left\| \hat{\boldsymbol{v}}_{\theta}(\boldsymbol{x}^{\{\text{outlet}\}}_{s}) - \boldsymbol{v}(\boldsymbol{x}^{\{\text{outlet}\}}_{s})\right\|^2, \label{eq:adapting:pinns:lossoutlet}\\
&\mathcal{L}_{\text{phy}}(\theta) = \frac{1}{P_{\text{phy}}} \sum_{r=1}^{N_{\text{phy}}} \left\| \hat{\boldsymbol{e}}_{\theta}(\boldsymbol{x}_r^{\{\text{phy}\}})\right\|^2 \label{eq:adapting:pinns:lossphy},
\end{align}
\end{subequations}
where $\left\| \cdot\right\|$ denotes the $L^2$ norm and the components of $\hat{\boldsymbol{e}}_{\theta} = (\hat{e}_{\theta 1},\hat{e}_{\theta 2},\hat{e}_{\theta 3},\hat{e}_{\theta 4})$  represent the residual of the governing NSE, 
\begin{subequations}
\begin{align}
	& \hat{e}_{\theta 1} = \rho_f(\hat{v}_{\theta 1}\frac{\partial \hat{v}_{\theta 1}}{\partial x_1} + \hat{v}_{\theta 2}\frac{\partial \hat{v}_{\theta 1}}{\partial x_2}+ \hat{v}_{\theta 3}\frac{\partial \hat{v}_{\theta 1}}{\partial x_3})  + \frac{\partial \hat{p}_{\theta}}{\partial x_1} - 
	\mu_f(\frac{\partial^2 \hat{v}_{\theta 1}}{\partial x_1^2} + \frac{\partial^2 \hat{v}_{\theta 1}}{\partial x_2^2} + \frac{\partial^2 \hat{v}_{\theta 1}}{\partial x_3^2}),\label{eq:adapting:pinns:loss:res1}\\
	& \hat{e}_{\theta 2} = \rho_f(\hat{v}_{\theta 1}\frac{\partial \hat{v}_{\theta 2}}{\partial x_1} + \hat{v}_{\theta 2}\frac{\partial \hat{v}_{\theta 2}}{\partial x_2}+ \hat{v}_{\theta 3}\frac{\partial \hat{v}_{\theta 2}}{\partial x_3}) + \frac{\partial \hat{p}_{\theta}}{\partial x_2} 
	- \mu_f(\frac{\partial^2 \hat{v}_{\theta 2}}{\partial x_1^2} + \frac{\partial^2 \hat{v}_{\theta 2}}{\partial x_2^2} + \frac{\partial^2 \hat{v}_{\theta 2}}{\partial x_3^2}),\label{eq:adapting:pinns:loss:res2}\\
	& \hat{e}_{\theta 3} = \rho_f(\hat{v}_{\theta 1}\frac{\partial \hat{v}_{\theta 3}}{\partial x_1} + \hat{v}_{\theta 2}\frac{\partial \hat{v}_{\theta 3}}{\partial x_2}+ \hat{v}_{\theta 3}\frac{\partial \hat{v}_{\theta 3}}{\partial x_3}) + \frac{\partial \hat{p}_{\theta}}{\partial x_3} - \mu_f(\frac{\partial^2 \hat{v}_{\theta 3}}{\partial x_1^2} + \frac{\partial^2 \hat{v}_{\theta 3}}{\partial x_2^2} + \frac{\partial^2 \hat{v}_{\theta 3}}{\partial x_3^2}),\label{eq:adapting:pinns:loss:res3}\\
	& \hat{e}_{\theta 4} = \frac{\partial \hat{v}_{\theta 1}}{\partial x_1} + \frac{\partial \hat{v}_{\theta 2}}{\partial x_2} + \frac{\partial \hat{v}_{\theta 3}}{\partial x_3}. \label{eq:adapting:pinns:loss:res4}
\end{align}
\end{subequations}

Notice we impose that the pressure is integrated into the learning process only through the loss data in Eq. \eqref{eq:adapting:pinns:lossdata} and the loss physics in Eq. \eqref{eq:adapting:pinns:lossphy} which corresponds to the training set structure described in Section \ref{sect:adapting:dataset}. In addition, the automatic differentiation process used to compute derivatives of the total loss function with respect to weights and biases, allows the computation of higher-order derivatives with respect to the input variables in Eq. \eqref{eq:adapting:pinns:lossphy}.   

\subsection{Techniques for Enhancing Physics-Informed Neural Networks}\label{sect:adapting:tech}

 While powerful, classical PINN often faces challenges such as spectral bias, where networks tend to prioritize the approximation of low frequencies over high frequencies (see \cite{Rahaman2019}), loss imbalance in the magnitude of the back-propagated gradients during model training using gradient descent (see \cite{Wang2021ModifiedMLP}) and optimization difficulties primarily due to the use of gradient-based optimization methods, for which the solution may get trapped in local minima. In the following, we will briefly discuss some of the best practices and techniques to mitigate these and other problems in PINNs and improve their performance while keeping the focus on the problem at hand. For more comprehensive information, we suggest \cite{Wang2023expertsguide}.
 
\subsubsection{Non-Dimensionalization}\label{sect:adapting:tech:dimensionless}
The Non-Dimensionalization technique is commonly used in science to simplify complex PDE systems by partially or completely eliminating physical dimensions through appropriate variable substitution. In particular, it involves transforming the original system, e.g. NSE  in Eqs. \eqref{eq:formulation:nse:momentum3D}-\eqref{eq:formulation:nse:bc:outlet} into an equivalent dimensionless form by choosing fundamental units or characteristic values, and scaling the variables in such a way they become dimensionless and typically of order one. This helps to prevent vanishing or exploding gradients in the neural networks and ensures that no single variable dominates the training process due to differences in scale, facilitating balanced learning of all variables. Recent studies that have used this technique include the following \cite{Gu2024,Jin2021,Garay2024}. In \ref{appendix::dimensionless}, we derive the dimensionless form of the AAA idealized model.

\subsubsection{Sampling within the Mesh}\label{sect:adapting:tech:sampling}
Adaptive sampling strategies can be employed to improve the training efficiency of PINNs. They focus on regions with high error or complex solution behavior, adjusting the sampling distribution based on model current predictions. In the literature, we can find examples of methods such as the residual-based adaptive refinement (RAR) introduced in \cite{Lu2021DeepXDE} and the Retain-Resample-Release sampling (R3) algorithm proposed by the authors in \cite{Daw2023}. 

As seen in Section \ref{sect:adapting:dataset}, in our work we rely on CFD simulations based on a meshing study adapted to the problem under study. Therefore, we take advantage of this prior knowledge in mesh generation and, in particular, we perform a random sampling within the mesh. Instead of using all available mesh points of the training dataset, we apply batch training, i.e. a random subset of collocation points is sampled at each iteration, reducing the computational load. 

\subsubsection{Optimizer and Learning Rate Scheduling} \label{sect:adapting:tech:learningrate}
Here, we use Adam optimizer \cite{Kingma2014Adam} due to its adaptive learning rates scheduler and robustness across a variety of problems. In this work, unless stated otherwise, the following parameters for Adam optimizer are used: beta$_1$ is set to 0.9, beta$_2$ to 0.999, and epsilon to 1.0e-8. An initial value for the learning rate $\alpha_0 = 0.001$ is considered, with an exponential decay schedule to gradually reduce the learning rate during training. In particular, the learning rate at step $t$  is given by $\alpha_t = \alpha_0 \cdot r^{\left(\frac{t}{s}\right)}$, where $r$ is the decay rate set to 0.95, $t$ represents the current step in the training process, and $s$ is the number of decay steps set to 3000. 

\subsubsection{Loss Balancing (Grad Norm)}\label{sect:adapting:tech:gradnorm}
One of the main challenges in training PINNs is managing multi-scale losses, such as the scale differences that may arise between $\mathcal{L}_{\text{data}}$, $\mathcal{L}_{\text{inlet}}, \mathcal{L}_{\text{outlet}}, \mathcal{L}_{\text{wall}}$ and $\mathcal{L}_{\text{phy}}$. A possible approach is to assign weights to each loss term during training $\lambda_{\text{data}}, \lambda_{\text{inlet}}, \lambda_{\text{outlet}}, \lambda_{\text{wall}}, \lambda_{\text{phy}}$, however manually choosing these weights is impractical since optimal values vary widely across different problems. So then, loss balancing techniques are aimed at automatically balancing multiple loss terms without the need for manual tuning of loss weights. In this work, we applied the Grad Norm weighting scheme presented in \cite{Wang2023expertsguide}, where the global weights $\hat{\lambda}_{i}$ are computed such that gradient norms of all weighted loss terms are equal, i.e., $\|\hat{\lambda}_{i} \nabla_\theta \mathcal{L}_{i}(\theta)\| = C $ for all $i \in [\text{data}, \text{inlet}, \text{wall}, \text{outlet}, \text{phy}]$, and $C$ is a constant. To stabilize training, weights are updated using a moving average, $\lambda_i^{(t+1)} = \beta \lambda^{(t)}_i + (1 - \beta) \hat{\lambda}_i^{(t+1)} $, where momentum $\beta$ determines the balance between the old and new values. This approach prevents any single loss term from dominating the training process, leading to improved stability and more consistent convergence. In this work, unless otherwise specified, Grad Norm is enabled using a momentum value of 0.9 and updates occurring every 1000 steps. 

\subsubsection{Fourier Features Embedding}\label{sect:adapting:tech:fourier}
 This technique addresses the spectral bias of neural networks towards low-frequency functions by enhancing the network's ability to learn high-frequency components. Specifically, it maps input coordinates into a higher-dimensional space via a random Fourier features embedding $\gamma:\mathbb{R}^{d}\to \mathbb{R}^{2e}$, defined as follows,
\begin{align}
   & \gamma(\boldsymbol{x}) = \left[\cos(2\pi\boldsymbol{B}\boldsymbol{x}), \sin(2\pi\boldsymbol{B}\boldsymbol{x}) \right]^{\text{T}} \nonumber
\end{align}
where  $\boldsymbol{B}\in \mathbb{R}^{e \times d}$ is a matrix with entries sampled from a Gaussian distribution $\mathcal{N}(0, \sigma^2)$.
Notice that the $e$ in the matrix dimensions represents the embedding dimension and, unless otherwise specified, it is set to 128. In addition,  a moderate value of $\sigma \in [1, 10]$ is suggested in the literature (see \cite{Tancik2020,Hennigh2021NVIDIA,Wang2021FourierF} for more details). Here, the standard deviation $\sigma$ is set to 1.0.

\subsubsection{Random Weight Factorization (RWF)}\label{sect:adapting:tech:rwf}
The aim of RWF, proposed by the authors in \cite{Wang2022RWF}, is to improve optimization by introducing trainable scale factors for each neuron, allowing the network to adjust weight magnitudes dynamically. For this purpose, it decomposes each neuron weight vector $\boldsymbol{w}^{(k,l)}$ into a product of a scalar $s^{(k,l)}$ and a vector $\boldsymbol{v}^{(k,l)}$, as follows,
 \begin{align}
    & \boldsymbol{w}^{(k,l)} = s^{(k,l)} \cdot \boldsymbol{v}^{(k,l)}. \nonumber
\end{align}

  The scale factor $s$ is initialized from $\mathcal{N}(\mu, \sigma^2)$, and then an exponential function is applied to ensure positivity, i.e., $s = \exp(\text{sampled value})$. 
Based on the literature, we choose the values for the mean and the standard deviation, unless otherwise specified, as follows $\mu = 0.5$ and $\sigma = 0.1$.
Finally, both $s$ and $\boldsymbol{v}$ are optimized using gradient descent.

\subsubsection{Modified Multi-Layer Perceptron (Modified-MLP)} \label{sect:adapting:tech:modifiedmlp}
 This technique, introduced in \cite{Wang2021ModifiedMLP}, increases the network capacity to learn nonlinear and complex PDE solutions by modifying the standard MLP architecture. The modifications involve the introduction of dual encoders, denoted as $\boldsymbol{U}=\sigma\left(\boldsymbol{W}_1\boldsymbol{x}+\boldsymbol{b}_1\right)$ and $\boldsymbol{V}=\sigma\left(\boldsymbol{W}_2\boldsymbol{x}+\boldsymbol{b}_2\right)$, which process input coordinates separately. Additionally, a layer-wise feature merging technique is implemented in each hidden layer of a standard MLP via a gating mechanism. The result is a weighted combination as follows,
  \begin{align}
    & \boldsymbol{g}^{(l)}(\boldsymbol{x}) = \sigma(\boldsymbol{f}^{(l)}(\boldsymbol{x})) \odot \boldsymbol{U} + (1 - \sigma(\boldsymbol{f}^{(l)}(\mathbf{x}))) \odot \boldsymbol{V}, \quad \text{for} \quad l = 1 ,\dots, L, \nonumber
\end{align}
 where $\boldsymbol{f}^{(l)}
(\boldsymbol{x})=\boldsymbol{W}^{(l)}\cdot\boldsymbol{g}^{(l-1)}(\boldsymbol{x})+\boldsymbol{b}^{(l)}$ represents the pre-activation output of the $l$-th layer in the MLP when given the input $\boldsymbol{x}$, $\sigma$ is a nonlinear activation function and $\odot$ denotes element-wise multiplication. The final network output is given by $\boldsymbol{f}_\theta(\boldsymbol{x})=\boldsymbol{W}^{(L+1)}\cdot\boldsymbol{g}^{(L)}(\boldsymbol{x})+\boldsymbol{b}^{(L+1)}$. Thus, all trainable parameters can be summarized as follows
\begin{align}
& \theta = \{\boldsymbol{W}_1, \boldsymbol{b}_1, \boldsymbol{W}_2, \boldsymbol{b}_2, (\boldsymbol{W}^{(l)}, \boldsymbol{b}^{(l)})_{l=1}^{L+1} \}. \nonumber
\end{align}

 The gating mechanism allows the network to improve its ability to capture complex patterns, resulting in more accurate solutions. On the downside, the modified architecture requires more computing resources due to the additional operations.

\subsection{DeepONet Architecture and Loss Functions}\label{sect:adapting:deeponet}

The vanilla (PI-)DeepONet described in Sections \ref{sect:background:deeponet} and \ref{sect:background:pideeponet} can be directly applied to predict velocities and pressure in AAA simulations (see, for instance, the ideas proposed in \cite{Jnini2024}). This involves training four separate (PI-)DeepONets to predict the three components of the velocity field and the pressure field, respectively. In each (PI-)DeepONet, the trunk net is a fully-connected neural network that receives the coordinate points 
$(x_1, x_2, x_3)$ as input. The branch net is also a fully-connected neural network that takes different inputs depending on the prediction target, i.e., it could receive the inlet velocity $v^{\{\text{inlet}\}}_{2}$ when predicting the velocity components and the outlet pressure $p^{\{\text{outlet}\}}$  when predicting the pressure field.

However, in this paper, we propose a more comprehensive architecture that integrates the results of papers \cite{Jin2022MultipleInput} and \cite{Wang2023Long} as extensions to vanilla (PI-)DeepONet, namely multiple inputs and multiple outputs. 

\subsubsection{Multiple Inputs}\label{sect:adapting:deeponet:mulinp}
An extension of (PI-)DeepONet is its ability to handle multiple inputs. While the vanilla architecture is designed for input functions on a single Banach space, many practical applications require processing several input functions simultaneously. To address this, the multiple input operators theorem was theoretically formulated and proposed in \cite{Jin2022MultipleInput}. The operator $\hat{\mathfrak{G}}_\theta$ for multiple input functions can be defined as follows,
{\small\begin{align}
	& \hat{\mathfrak{G}}_\theta(f^{(i)}_1, f^{(i)}_2,\dots,f^{(i)}_n)(\boldsymbol{x}^{(i)}_j, t^{(i)}_j):=\sum_{k=1}^q\underset{Branch_1}{\underbrace{{\beta}^{(1)}_{k}\left(\varphi^{(1)}_{m_1}(f^{(i)}_1)\right)}}\;\underset{Branch_2}{\underbrace{{\beta}^{(2)}_{k}\left(\varphi^{(2)}_{m_2}(f^{(i)}_2)\right)}}\cdots\underset{Branch_n}{\underbrace{{\beta}^{(n)}_{k}\left(\varphi^{(n)}_{m_n}(f^{(i)}_n)\right)}}\;\underset{Trunk}{\underbrace{\tau_k(\boldsymbol{x}^{(i)}_j, t^{(i)}_j)}}, \label{eq:adapting:deeponet:definition:multipleinput}
\end{align}}
where
\begin{align}
	& \varphi^{(\xi)}_{m_\xi}(f^{(i)}_\xi) = \left(f_\xi^{(i)}(\tilde{\boldsymbol{x}}^{(\xi)}_1),f_\xi^{(i)}(\tilde{\boldsymbol{x}}^{(\xi)}_2),\dots,f_\xi^{(i)}(\tilde{\boldsymbol{x}}^{(\xi)}_{m_\xi})\right), \quad \xi = 1,\dots,n. \label{eq:adapting:deeponet:definition:phi}
\end{align}

 Here, the architecture has $n$ independent branch nets and one trunk net. The $n$-th branch net encodes the input function $f^{(i)}_n$ via $\varphi^{(n)}_{m_n}(f^{(i)}_n)$ and outputs $\left\{\beta_k^{(n)}\right \}_{k=1}^{q}$, while the trunk net encodes $(\boldsymbol{x}^{(i)}_j, t^{(i)}_j)$ and outputs $\left\{\tau_k\right \}_{k=1}^{q}$. This architecture is known in the literature as MIONet-low-rank (see \cite{Jin2022MultipleInput}) and the connection to the vanilla (PI-)DeepONet occurs when only one input function is considered, i.e., n = 1 (see Eq. \eqref{eq:background:deeponet:definition}). 

Notice that the definition for $\varphi^{(\xi)}_{m_\xi}(f^{(i)}_\xi)$ in Eq. \eqref{eq:adapting:deeponet:definition:phi} assume the Faber-Schauder basis in \( C[0,1] \), so then, it is computed by evaluating the function at the first \( m_\xi \) grid points and forming a vector with these values. For a more general definition, please refer to  \cite{Jin2022MultipleInput}.

\subsubsection{Multiple Outputs}\label{sect:adapting:deeponet:mulout}
As observed in Figure \ref{fig:background:deeponet}, the output of a vanilla (PI-)DeepONet is a scalar, while the solution of the NSE in Eqs. \eqref{eq:formulation:nse:momentum3D}-\eqref{eq:formulation:nse:bc:outlet} comprises the 
three components of the velocity vector field and the pressure scalar field. Thus, in this case, we need a (PI-)DeepONet which returns a vector-value function. The authors in \cite{Wang2023Long} modified the original forward pass of the method as follows, 
\begin{align}
	& \hat{\mathfrak{G}}^{(\zeta)}_\theta(f^{(i)})(\boldsymbol{x}^{(i)}_j, t^{(i)}_j):=\sum_{k=q_{\zeta-1} + 1}^{q_\zeta}\underset{Branch}{\underbrace{{\beta}_{k}\left(f^{(i)}(\tilde{\boldsymbol{x}}_1),f^{(i)}(\tilde{\boldsymbol{x}}_2),...,f^{(i)}(\tilde{\boldsymbol{x}}_m)\right)}}\;\underset{Trunk}{\underbrace{\tau_k(\boldsymbol{x}^{(i)}_j, t^{(i)}_j)}}, \quad \zeta = 1, \dots, \check{n}, \label{eq:adapting:deeponet:definition:multipleoutput}
\end{align}
where $0= q_0 < q_1 < \cdots < q_{\check{n}} = q$. This simple modification enables the (PI-)DeepONet to output an $\check{n}$-dimensional vector, that is $\hat{\mathfrak{G}}_\theta = [\hat{\mathfrak{G}}^{(1)}_\theta, \hat{\mathfrak{G}}^{(2)}_\theta, \dots, \hat{\mathfrak{G}}^{(\check{n})}_\theta]$.

\subsubsection{Proposed Architecture.}\label{sect:adapting:mydeeponet}

\begin{figure}[htbp]
	\centering	
	\includegraphics[scale=0.9]{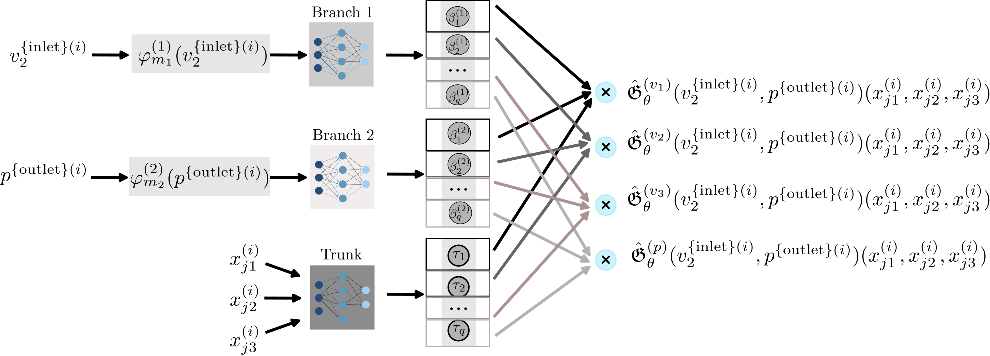}
	\caption{MI-MO-DeepONet architecture.}
	\label{fig:adapting:mydeeponet}
\end{figure}

Based on the AAA idealized model, we set $n = 2$ (see Eq. \eqref{eq:adapting:deeponet:definition:multipleinput}), resulting in an architecture with two branch networks: one receiving the inlet velocity $v^{\{\text{inlet}\}}_{2}$ and the other receiving the outlet pressure $p^{\{\text{outlet}\}}$. Additionally, we take $\check{n} = 4$ (see Eq. \eqref{eq:adapting:deeponet:definition:multipleoutput}) enabling a $4$-dimensional vector as output. Finally, the proposed architecture can be seen in Figure \ref{fig:adapting:mydeeponet}.
 
\subsubsection{Loss Functions}\label{sect:adapting:deeponet:loss}

We adopt a similar approach as in Section \ref{sect:adapting:pinns} for PINNs. In particular, we differentiate between the total loss function for a standard DeepONet, which relies solely on known solution data through $\mathcal{L}_{\text{data}}$ and incorporates boundary conditions via $\mathcal{L}_{\text{inlet}}, \mathcal{L}_{\text{outlet}}, \mathcal{L}_{\text{wall}}$, representing a purely data-driven approach, and the total loss function for PI-DeepONet, which additionally integrates the Navier-Stokes equations through $\mathcal{L}_{\text{phy}}$. That is,
\begin{subequations}
\begin{align}
    & \mathcal{L}_{\text{DeepONet}}(\theta) = \mathcal{L}_{\text{data}}(\theta) +  \mathcal{L}_{\text{inlet}}(\theta) +  \mathcal{L}_{\text{outlet}}(\theta) + \mathcal{L}_{\text{wall}}(\theta), \label{eq:adapting:deeponet:loss}\\
    & \mathcal{L}_{\text{PI-DeepONet}}(\theta) = \mathcal{L}_{\text{data}}(\theta) +  \mathcal{L}_{\text{inlet}}(\theta) + \mathcal{L}_{\text{outlet}}(\theta) +  \mathcal{L}_{\text{wall}}(\theta) + \mathcal{L}_{\text{phy}}(\theta). \label{eq:adapting:pideeponet:loss}
\end{align}
\end{subequations}

Now, the loss components are defined as follows, 
\begin{subequations}
\begin{align}
& \mathcal{L}_{\text{data}}(\theta) = \frac{1}{N P_{\text{data}}}\sum_{i=1}^{N}\sum_{j=1}^{P_{\text{data}}} \left( \left\| \hat{\mathfrak{G}}^{(\boldsymbol{v})}_\theta(v^{\{\text{inlet}\}(i)}_2,p^{\{\text{outlet}\}(i)})(\boldsymbol{x}^{\{\text{data}\}(i)}_{j}) - \boldsymbol{v}(\boldsymbol{x}^{\{\text{data}\}(i)}_{j})\right\|^2 \right. \nonumber \\
& + \left. \left| \hat{\mathfrak{G}}^{(p)}_\theta(v^{\{\text{inlet}\}(i)}_2,p^{\{\text{outlet}\}(i)})(\boldsymbol{x}^{\{\text{data}\}(i)}_{j}) - p(\boldsymbol{x}^{\{\text{data}\}(i)}_{j})\right|^2 \right),\label{eq:adapting:pideeponet:lossdata}\\
&\mathcal{L}_{\text{inlet}}(\theta) = \frac{1}{N P_{\text{inlet}}} \sum_{i=1}^{N}\sum_{k=1}^{P_{\text{inlet}}} \left\| \hat{\mathfrak{G}}^{(\boldsymbol{v})}_\theta(v^{\{\text{inlet}\}(i)}_2,p^{\{\text{outlet}\}(i)})(\boldsymbol{x}^{\{\text{inlet}\}(i)}_{k}) - \boldsymbol{v}(\boldsymbol{x}^{\{\text{inlet}\}(i)}_{k})\right\|^2,\label{eq:adapting:pideeponet:lossinlet}\\ 
&\mathcal{L}_{\text{wall}}(\theta) = \frac{1}{NP_{\text{wall}}}\sum_{i=1}^{N}\sum_{l=1}^{P_{\text{wall}}} \left\| \hat{\mathfrak{G}}^{(\boldsymbol{v})}_\theta(v^{\{\text{inlet}\}(i)}_2,p^{\{\text{outlet}\}(i)})(\boldsymbol{x}^{\{\text{wall}\}(i)}_{l}) - \boldsymbol{v}(\boldsymbol{x}^{\{\text{wall}\}(i)}_{l})\right\|^2, \label{eq:adapting:pideeponet:losswall}\\  
&\mathcal{L}_{\text{outlet}}(\theta) = \frac{1}{N P_{\text{outlet}}}\sum_{i=1}^{N}\sum_{s=1}^{P_{\text{outlet}}} \left\| \hat{\mathfrak{G}}^{(\boldsymbol{v})}_\theta(v^{\{\text{inlet}\}(i)}_2,p^{\{\text{outlet}\}(i)})(\boldsymbol{x}^{\{\text{outlet}\}(i)}_{s}) - \boldsymbol{v}(\boldsymbol{x}^{\{\text{outlet}\}(i)}_{s})\right\|^2, \label{eq:adapting:pideeponet:lossoutlet}\\
&\mathcal{L}_{\text{phy}}(\theta) = \frac{1}{N P_{\text{phy}}} \sum_{i=1}^{N}\sum_{r=1}^{N_{\text{phy}}} \left\| \hat{\boldsymbol{E}}_{\theta}(v^{\{\text{inlet}\}(i)}_2,p^{\{\text{outlet}\}(i)})(\boldsymbol{x}_r^{\{\text{phy}\}(i)})\right\|^2 \label{eq:adapting:pideeponet:lossphy},
\end{align}
\end{subequations}
where  the components of $\hat{\boldsymbol{E}}_{\theta} = (\hat{E}_{\theta 1},\hat{E}_{\theta 2},\hat{E}_{\theta 3},\hat{E}_{\theta 4})$  represent the residual of the governing NSE, 
\begin{subequations}
{\small\begin{align}
	& \hat{E}_{\theta 1} = \rho_f(\hat{\mathfrak{G}}^{(v_1)}_\theta\frac{\partial \hat{\mathfrak{G}}^{(v_1)}_\theta}{\partial x_1} + \hat{\mathfrak{G}}^{(v_2)}_\theta\frac{\partial \hat{\mathfrak{G}}^{(v_1)}_\theta}{\partial x_2}+ \hat{\mathfrak{G}}^{(v_3)}_\theta\frac{\partial \hat{\mathfrak{G}}^{(v_1)}_\theta}{\partial x_3})  + \frac{\partial \hat{\mathfrak{G}}^{(p)}_\theta}{\partial x_1} - 
	\mu_f(\frac{\partial^2 \hat{\mathfrak{G}}^{(v_1)}_\theta}{\partial x_1^2} + \frac{\partial^2 \hat{\mathfrak{G}}^{(v_1)}_\theta}{\partial x_2^2} + \frac{\partial^2 \hat{\mathfrak{G}}^{(v_1)}_\theta}{\partial x_3^2}),\label{eq:adapting:pideeponet:lossphy:res1}\\
	& \hat{E}_{\theta 2} = \rho_f(\hat{\mathfrak{G}}^{(v_1)}_\theta\frac{\partial \hat{\mathfrak{G}}^{(v_2)}_\theta}{\partial x_1} + \hat{\mathfrak{G}}^{(v_2)}_\theta\frac{\partial \hat{\mathfrak{G}}^{(v_2)}_\theta}{\partial x_2}+ \hat{\mathfrak{G}}^{(v_3)}_\theta\frac{\partial \hat{\mathfrak{G}}^{(v_2)}_\theta}{\partial x_3}) + \frac{\partial \hat{\mathfrak{G}}^{(p)}_\theta}{\partial x_2} 
	- \mu_f(\frac{\partial^2 \hat{\mathfrak{G}}^{(v_2)}_\theta}{\partial x_1^2} + \frac{\partial^2 \hat{\mathfrak{G}}^{(v_2)}_\theta}{\partial x_2^2} + \frac{\partial^2 \hat{\mathfrak{G}}^{(v_2)}_\theta}{\partial x_3^2}),\label{eq:adapting:pideeponet:lossphy:res2}\\
	& \hat{E}_{\theta 3} = \rho_f(\hat{\mathfrak{G}}^{(v_1)}_\theta\frac{\partial \hat{\mathfrak{G}}^{(v_3)}_\theta}{\partial x_1} + \hat{\mathfrak{G}}^{(v_2)}_\theta\frac{\partial \hat{\mathfrak{G}}^{(v_3)}_\theta}{\partial x_2}+ \hat{\mathfrak{G}}^{(v_3)}_\theta\frac{\partial \hat{\mathfrak{G}}^{(v_3)}_\theta}{\partial x_3}) + \frac{\partial \hat{\mathfrak{G}}^{(p)}_\theta}{\partial x_3} - \mu_f(\frac{\partial^2 \hat{\mathfrak{G}}^{(v_3)}_\theta}{\partial x_1^2} + \frac{\partial^2 \hat{\mathfrak{G}}^{(v_3)}_\theta}{\partial x_2^2} + \frac{\partial^2 \hat{\mathfrak{G}}^{(v_3)}_\theta}{\partial x_3^2}),\label{eq:adapting:pideeponet:lossphy:res3}\\
	& \hat{E}_{\theta 4} = \frac{\partial \hat{\mathfrak{G}}^{(v_1)}_\theta}{\partial x_1} + \frac{\partial \hat{\mathfrak{G}}^{(v_2)}_\theta}{\partial x_2} + \frac{\partial \hat{\mathfrak{G}}^{(v_3)}_\theta}{\partial x_3}. \label{eq:adapting:pideeponet:lossphy:res4}
\end{align}}
\end{subequations}

\section{Results}\label{sect:results}
In this section, we conduct several numerical experiments to evaluate the performance of PINN and (PI-)DeepONet models in predicting velocity and pressure fields in the context of AAA simulations. The models were developed using the PyTorch framework, with computations being performed on a locally hosted PC featuring an NVIDIA RTX A6000 GPU with 48 GB of memory. To assess performance, we use the $L^2$-relative error, which measures the discrepancy between the solutions predicted by the neural networks and the ground trust solutions. Furthermore, the relative error reported is the mean of the $L^2$-relative errors calculated using a batch size of 10 000 coordinate points.

It is worth mentioning that due to the computation time required for the training phase in each case, we have manually adjusted the hyper-parameters based on commonly used values found in the literature. Attempting to determine the absolute optimal hyper-parameter settings is beyond the scope of this paper and will be addressed in future work. However, the selected hyper-parameters yield results with a satisfactory level of accuracy and effectively demonstrate the outlined goals.

\subsection{AAA Simulations via PINNs}\label{sect:results:pinns}

\subsubsection{DeepNNs Vs PINNs}\label{sect:results:pinns:drf}

The purpose of this section is to compare DeepNNs and PINNs, focusing on the impact of incorporating the governing physics equations into the loss function (see Eqs.\eqref{eq:adapting:deepnn:loss}-\eqref{eq:adapting:pinns:loss}). Additionally, we study how the resolution and data formatting (see Figure \ref{fig:adapting:datascenarios}) affect the models' predictions. In this particular case, we set the maximum inlet velocity $V$ to 0.1 $m/s$.

Regarding the models' settings, we use a neural network architecture consisting of a MLP with 4 hidden layers and 256 neurons per layer, using the Tanh activation function. Models are trained for 200 000 iterations with a batch size of 1024, employing batch training. The parameters are initialized using Xavier normal initialization \cite{Glorot2010}, and the Adam optimizer is utilized for optimization. In addition, from Section \ref{sect:adapting:tech}, we use dimensionless Navier-Stokes equations (NSE), apply the sampling within the mesh, and implement exponential decay for the learning rate combined with an optimizer scheduler. We also employ loss balancing through the Grad Norm technique to further improve models' performance.

\begin{table}[h!]
		\centering
	\resizebox{\columnwidth}{!}{	
		\begin{tabular}{||l|c||c|c|c||c|c|c||c|c|c||}
			\hline
			\multicolumn{2}{||c||}{\textbf{Dataset}}  & \multicolumn{3}{|c||}{\textbf{DeepNN}}  & \multicolumn{3}{|c||}{\textbf{PINN}} & \multicolumn{3}{|c||}{\textbf{WU-PINN}}  \\
			\multicolumn{2}{||c||}{$V = 0.1\; m/s$ } & \multicolumn{2}{|c}{ $L^2$-relative error} & \multicolumn{1}{c||}{}  & \multicolumn{2}{|c}{ $L^2$-relative error} & \multicolumn{1}{c||}{}  & \multicolumn{2}{|c}{ $L^2$-relative error} & \multicolumn{1}{c||}{}  \\ 
			\hline
			Source of data  &  \%  &  \rotatebox{90}{Magnitude of velocity} &  \rotatebox{90}{Pressure}
			&  \rotatebox{90}{Runtime(min)} &  \rotatebox{90}{Magnitude of velocity} &  \rotatebox{90}{Pressure}
			&  \rotatebox{90}{Runtime(h)} &  \rotatebox{90}{Magnitude of velocity} &  \rotatebox{90}{Pressure}
			&  \rotatebox{90}{Runtime(h)} \\ \hline
			Cross-section &    &   &  
			&  &   &  
			&   &   & 
			&   \\
			5 slices & 1.42 
   & 1.1881e-01 
   &  1.5626e-01 
   & 28.30 
   & 4.6507e-02 
   & 4.8523e-02 
   & 2.95
   & 6.1912e-02 
   & 6.9856e-02 
   & 1.17\\
			\hline
			Longitudinal &  &  &  &  &  &  &  &  &  &\\
			1 slice 
   & 1.87 
   &   2.7889e-01
   &  7.0684e-01 
   & 28.11  
   & 4.0500e-02 
   &  1.1026e-02 
   & 2.95  
   &  6.7602e-02 
   & 1.6320e-02 
   & 1.17\\
			\hline
			Random &  &  &  &  &  &  &  &  &  & \\
			Coord. points & 0.3 
   &  2.3253e-02 
   &  1.4570e-02 
   & 28.84 
   & 8.7983e-03 
   & 4.7490e-03 
   & 2.95  
   & 1.3837e-02 
   & 7.9948e-03
			& 1.15 \\
			\hline
		\end{tabular} }
		\caption{ Comparison between DeepNN, PINN, and WU-PINN models.}
        \label{tab:results:pinns:drf}
	\end{table}

In Table \ref{tab:results:pinns:drf}, we show the results of the comparison between DeepNN and PINN models in terms of performance and computational efficiency. In particular, we report the $L^2$-relative error on the test dataset, measuring the discrepancy between the predicted and ground truth values for both the magnitude of velocity, defined as $|\boldsymbol{v}(\boldsymbol{x})| = \sqrt{v_1^2(\boldsymbol{x})+v_2^2(\boldsymbol{x})+v_3^2(\boldsymbol{x})}$, and for the pressure. Additionally, we include the runtime required for training each of these models. Notice that the symbol \% indicates the percentage that Data represent of the Volume region in the training set for the three scenarios, i.e., `cross-section', `longitudinal', and `random', respectively.

As observed, PINN outperforms DeepNN in accuracy for all three cases, with the largest discrepancy being exhibited in the `longitudinal' case. The third case shows the lowest relative errors (best results) and also the smallest difference between the methods' results. In contrast, training the DeepNN model is $5x$ faster in terms of computational efficiency compared to the PINN models. This speed advantage is because DeepNN does not include the physical loss term $\mathcal{L}_{\text{phy}}$, avoiding the computationally expensive process of automatic differentiation.

 For the sake of improving the computation time of the training phase in the PINN model, while maintaining high accuracy, we take inspiration from the work of \cite{Daneker2024} and propose a Warm-Up Physics-Informed Neural Networks (WU-PINNs) to be included in the comparison (see Table \ref{tab:results:pinns:drf}). The WU-PINN involves training the DeepNN as a "warm-up" phase, followed by transfer learning to the PINN model that incorporates the loss physics. In particular, after fixing all previous hyper-parameters, we first train the DeepNN for 150 000 iterations, saving the best-performing model. We then perform transfer learning, using this pre-trained model to initialize the PINN, and train it for 50 000 iterations, matching the total 200 000 iterations of the previous models. As shown in the table the results significantly improved in accuracy with respect to DeepNN and at the same time reduced the training runtime by half with respect to PINN. Note that the training runtime of WU-PINN is the combined time of training DeepNN and PINN models for 150 000 and 50 000 iterations, respectively.
 
\subsubsection{Processing Noisy Data}\label{sect:results:pinns:noise}
\begin{figure}[htbp]
		\centering	
		\resizebox{\columnwidth}{!}{
  \includegraphics[scale=0.035]{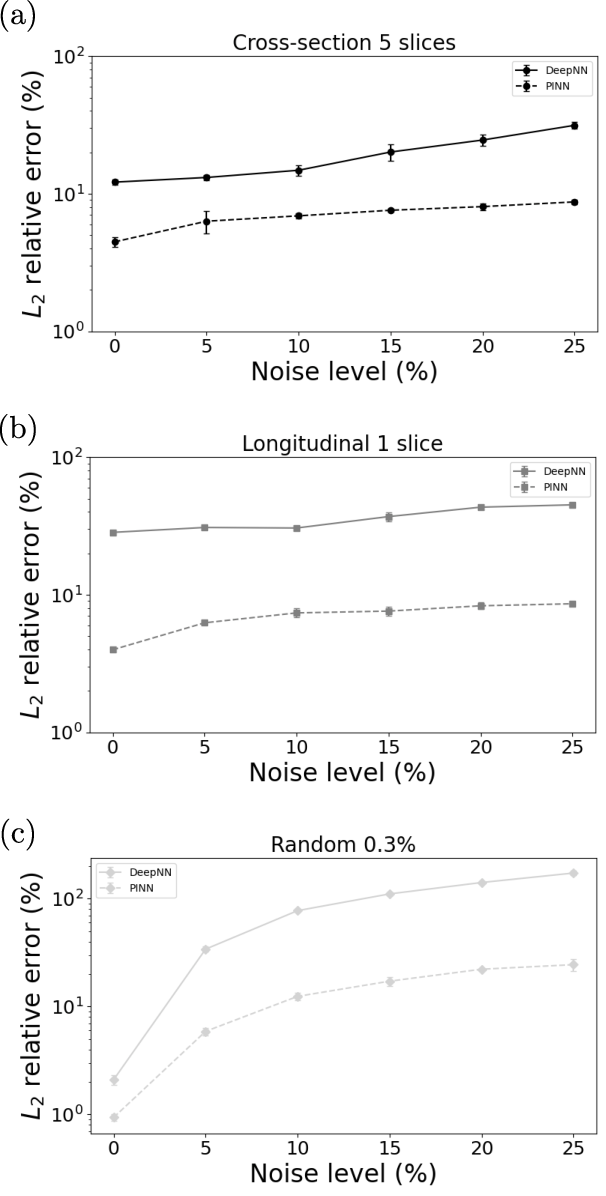}
  }
		\caption{Analysis of the impact of noise via the $L^2$-relative errors (\%) for the velocity magnitude. Comparison between DeepNN and PINN. }
		\label{fig:results:pinns:noise}
	\end{figure}

Now, we aim to study how PINN performs when handling noisy data, a scenario particularly relevant in clinical settings where measurement noise is common. Similar to the previous section, we set the maximum inlet velocity $V$ to 0.1 $m/s$. 

To simulate this inherent noise in our external mimicking source of known Data (see Figure \ref{fig:adapting:datascenarios}), we introduce Gaussian noise by generating random values with a mean of zero and a standard deviation (SD) denoted by a factor $\sigma$, matching the shape of Data. The SD is proportional to the highest velocity value in the full dataset, calculated as $\sigma$ = noise level $\times$ max velocity.  The parameter 'noise level' varies from 0\% to 25\%, representing different levels of noise intensity. The generated noise is then added to the velocity field in the Data, and this noisy version is used during training. It is worth mentioning that during training, data from the boundary regions of the domain (Inlet$^*$, Outlet$^*$, Wall$^*$) are not affected by the noise nor are the validation or test datasets.

In Figure \ref{fig:results:pinns:noise}, we present the impact of noise via the $L^2$-relative errors (\%) on the test dataset for the velocity magnitude only. Different noise levels for the three distinct data formatting scenarios (see Figure \ref{fig:adapting:datascenarios}) are given. The computations are performed three times by means of different random seeds to account for prediction variability, with error bars representing standard deviations from these three trials. For each scenario, we compare the performance of the DeepNN and PINN models, using the same architecture, hyper-parameters and techniques as in Section \ref{sect:results:pinns:drf}. At 0\% noise, the results reproduce those in Table \ref{tab:results:pinns:drf} as a particular case. As noise levels increase, the error growth varies depending on the data format, with the first two scenarios showing higher robustness to noise. Overall, PINN exhibits a remarkable resilience to noise, consistently outperforming the data-driven DeepNN model under these conditions.
\subsubsection{PINNs without Data}\label{sect:results:pinns:pnd}
In this section, we explore an extreme case of PINNs that operate without any data, other than the boundary conditions and the unsupervised constraints imposed by the Navier-Stokes equations (NSE). This is achieved by removing the data loss component $\mathcal{L}_{\text{data}}$ from Eq. \eqref{eq:adapting:pinns:loss}. Such a setup is particularly valuable for solving problems where experimental data or numerical simulations may not be available, allowing the model to be guided entirely by the underlying physics of the system. Note that, under these conditions, a DeepNN model would not have enough data to be properly trained.

\begin{table}[h!]
	\centering
	\resizebox{\columnwidth}{!}{
		\begin{tabular}{|c|c|c|c|c|c|c|c|}
		\hline
		\multicolumn{5}{|c|}{300 000 iter \quad - \quad batch size 3000} &  \multicolumn{2}{c}{$L^2$-relative error} &  \multicolumn{1}{|c|}{} \\
		\hline
		\textbf{Exponential decay} & \textbf{Grad Norm} & \textbf{Fourier embedding} & \textbf{Modified MLP} & \textbf{RWF} & \textbf{Magnitude of velocity} & \textbf{Pressure} & \textbf{Runtime (h)}\\
		\hline
		\textcolor{gray}{\checkmark} & \textcolor{gray}{\checkmark} & \textcolor{gray}{\checkmark} & \textcolor{gray}{\checkmark} & \textcolor{gray}{\checkmark} & 2.4912e-02
 & 2.8641e-02
 & 9.60
 \\
		\hline
		\textcolor{gray}{\checkmark} & \textcolor{gray}{\checkmark} & \textcolor{gray}{\checkmark} & \textcolor{gray}{\checkmark} & \textbf{\textcolor{black}{\texttimes}} & 2.8433e-02
& 4.4519e-02
 & 9.57
 \\
		\hline
		\textbf{\textcolor{black}{\texttimes}} & \textcolor{gray}{\checkmark} & \textcolor{gray}{\checkmark} & \textcolor{gray}{\checkmark} & \textcolor{gray}{\checkmark} & 3.6517e-02
 & 4.7310e-02
 & 9.62
\\
		\hline
		\textcolor{gray}{\checkmark} & \textcolor{gray}{\checkmark} & \textbf{\textcolor{black}{\texttimes}} & \textcolor{gray}{\checkmark} & \textcolor{gray}{\checkmark} & 4.0986e-02
 & 3.0455e-02
 & 8.37
\\
		\hline
		\textcolor{gray}{\checkmark} & \textbf{\textcolor{black}{\texttimes}} & \textcolor{gray}{\checkmark} & \textcolor{gray}{\checkmark} & \textcolor{gray}{\checkmark} & 4.3730e-02
 & 4.1168e-02
 & 5.51
 \\
		\hline

		\textcolor{gray}{\checkmark} & \textcolor{gray}{\checkmark} & \textcolor{gray}{\checkmark} & \textbf{\textcolor{black}{\texttimes}} & \textcolor{gray}{\checkmark} & 6.1967e-02
 & 1.1688e-01
  & 5.43
\\
		\hline
		\textbf{\textcolor{black}{\texttimes}} & \textbf{\textcolor{black}{\texttimes}} & \textbf{\textcolor{black}{\texttimes}} & \textbf{\textcolor{black}{\texttimes}} & \textbf{\textcolor{black}{\texttimes}} & 2.2596e-01

 & 5.0376e-01

  & 4.21

\\
		\hline

	\end{tabular}}
	\caption{Ablation study}
 \label{tab:results:pinns:pnd:ablation}
\end{table}

Here, the maximum inlet velocity $V$ is set to 0.1 $m/s$. The PINN model uses the same neural network architecture, hyper-parameters and techniques as in Section \ref{sect:results:pinns:drf}, with the difference that this time the training phase performs 300 000 iterations with a batch size of 3000. In addition, to further enhance the model's ability to capture complex patterns and optimize its learning process, we incorporate Fourier feature embedding and RWF techniques, and the Modified-MLP architecture (see Section \ref{sect:adapting:tech}).

In Table \ref{tab:results:pinns:pnd:ablation}, we present the $L^2$-relative errors on the test dataset. The results are analyzed through an Ablation study, which demonstrates the impact of each implemented technique by progressively deactivating them, showing their influence on the accuracy of the model. As expected, the most accurate result is obtained when all techniques are activated. In addition, the technique that contributes the most to minimizing the relative error is the Modified MLP. 

It is worth mentioning that because we remove the data loss component, the pressure is integrated into the learning process only through
the loss physics (see Eq. \eqref{eq:adapting:pinns:lossphy}). In this framework, it occurs that the model constantly overestimates or underestimates the pressure by a quasi-constant shift, which we define as the mean value of the point shifts. This value is used to adjust the final pressure predicted by the model (see Table \ref{tab:results:pinns:pnd:ablation}). 

\begin{figure}[ht]
	\centering	
	\resizebox{\columnwidth}{!}{\includegraphics[scale=0.5]{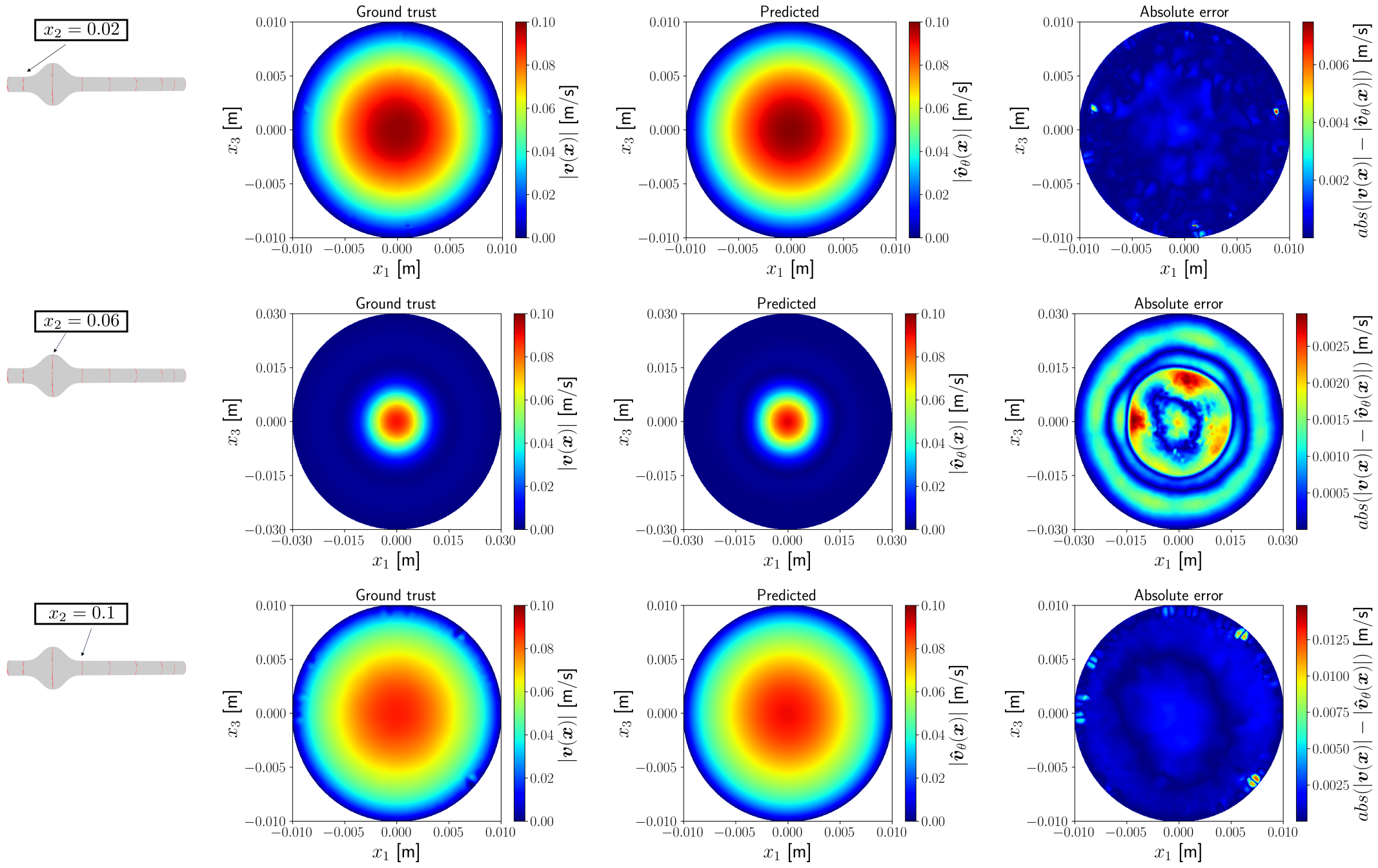}}
	\caption{Visual representation of the magnitude of velocity in $m/s$ across three cross-section planes. Comparison of ground trust and predicted values using the absolute error.}
	\label{fig:results:pinns:pnd:cross-section}
\end{figure}

Finally, considering the case where all techniques are activated, Figure \ref{fig:results:pinns:pnd:cross-section} shows a comparison between the ground trust and the predicted values of the magnitude of the velocity via a visual representation of the absolute error across different relevant cross-section planes. This highlights the effectiveness of the applied techniques to accurately capture the velocity distribution.

\subsubsection{Transfer Learning for Accelerating the Training Process}\label{sect:results:pinns:tl}
In Section \ref{sect:results:pinns:drf}, we introduced WU-PINN, a methodology designed to accelerate PINN using a previous DeepNN training phase. However, in this section, we want to apply a transfer learning approach to study the acceleration of the training phase in PINNs for the data-less scenario studied in Section \ref{sect:results:pinns:pnd}, where no data is available to pre-train a DeepNN model. Here, we reuse the architecture and hyper-parameters of Section \ref{sect:results:pinns:pnd}, and assume that all techniques are activated (see Table \ref{tab:results:pinns:pnd:ablation}).

To begin with, we first train benchmark PINN models from scratch, targeting different maximum inlet velocities such as $V_{\text{target}} = [0.04, 0.05, 0.06, 0.08, 0.12, 0.13, 0.15]m/s$, respectively, and capturing in each case the convergence data (see PINN in Table \ref{tab:results:pinns:tl}). In particular, we record the best iteration, which is the iteration in which the minimum total loss was reached, we also store the minimum total loss, training runtime and $L^2$-relative error on the test datasets for the magnitude of velocity and for the pressure. 

We continue by training a baseline model with a maximum inlet velocity of 0.1 $m/s$ that will serve as the foundation for subsequent models (as observed, such a model has already been trained in Section \ref{sect:results:pinns:pnd}). Then, we implement transfer learning by initializing a new model for each maximum inlet velocity in $V_{\text{target}}$ based on the optimal weights of the baseline model, and train them. The stopping criterion is now set at the minimum total loss obtained in the benchmark-trained counterparts, which provides a consistent reference point for the comparison.

In Table \ref{tab:results:pinns:tl}, we present the performance gains offered by the transfer learning approach for this particular case. By comparing the `best iter' in PINN and the `stop iter' in PINN\_TL, results suggest that the baseline model provides a more informed starting point for the training process, which leads to reaching the set minimum total loss more quickly. This is reflected in an improvement of the computational efficiency of the training phase as shown in the `time reduction'. The strategy leverages transfer learning as an effective means to accelerate PINN training in the data-free scenario, highlighting the adaptability and generalizability of PINNs across nearby physical settings. 

\begin{table}[!ht] 
	\centering
	\resizebox{\columnwidth}{!}{
\begin{tabular}{|c|c|c|c|c|c|c|c|} 
\hline
\multicolumn{1}{|c|}{\textbf{PINN}} & \textbf{V004} & \textbf{V005} & \textbf{V006} & \textbf{V008} & \textbf{V012} & \textbf{V013} & \textbf{V015} \\ 
\hline
best iter/total iter 
& 280347/300000
& 298351/300000
 & 299484/300000
 & 299491/300000
 & 294226/300000
 & 298296/300000
& 298021/300000\\ 
\hline
total loss  
& 1.3268e-05
 &9.9654e-06
 &8.6750e-06
 &6.8742e-06
 &5.0927e-06
 &4.7613e-06
 &4.6567e-06
\\ 
\hline
runtime (h) 
& 9.67
 &9.63
 &9.71
 &9.69
&9.72
 &9.61
 &9.67
 \\ 
\hline
vel l2\_error 
& 8.5579e-03
 &1.04884e-02
 &1.2793e-02
 &1.9020e-02
 &3.0042e-02
 &3.1872e-02
 &3.6203e-02
 \\ 
\hline
pressure l2\_error 
& 1.4034e-02
 &1.7318e-02
 &1.9759e-02
 &2.4772e-02
 &3.0369e-02
 &2.9095e-02
 &3.0574e-02
 \\ 
\hline
\multicolumn{1}{|c|}{\textbf{PINN\_TL}} & \textbf{V001 $\rightarrow$ V004} & \textbf{V001 $\rightarrow$ V005} & \textbf{V001 $\rightarrow$ V006} & \textbf{V001 $\rightarrow$ V008} & \textbf{V001 $\rightarrow$ V012} & \textbf{V001 $\rightarrow$ V013} & \textbf{V001 $\rightarrow$ V015} \\ 
\hline
stop iter/total iter 
& 207023/300000
 & 225147/300000
 & 217960/300000
 & 210344/300000
 & 197818/300000
 & 193030/300000
 & 183438/300000
 \\ 
\hline
runtime (h) 
& 6.66
 & 7.25
 & 7.00
 & 6.76
& 6.37
& 6.22
& 5.91
 \\ 
\hline
time reduction (h)
& 3.01
 & 2.38
& 2.71
 & 2.93
& 3.35
 & 3.39
 & 3.77
 \\ 
\hline
vel l2\_error 
& 1.0026e-02
 & 1.2309e-02
 & 1.4808e-02
 & 1.9014e-02
 & 2.6247e-02
& 2.7718e-02
 & 2.9891e-02
 \\ 
\hline
pressure l2\_error 
& 1.3393e-02
& 1.5689e-02
& 1.8484e-02
 & 2.3146e-02
 & 2.8122e-02
 & 2.8990e-02
 & 2.9337e-02
 \\ 
\hline
\end{tabular} }
 \caption{Transfer learning results.}
 \label{tab:results:pinns:tl}
\end{table}

Note that using the minimum total loss obtained in the benchmark trained counterparts as stopping conditions is practical for our study, but the transferred models could achieve better minima if we let them train for 300 000 iterations due to the informed initialization. In addition, it is crucial to ensure the consistency of the network architectures in all experiments to be compatible when transferring weights. 

\subsection{AAA Simulations via (PI-)DeepONet}\label{sect:results:deeponet}

\subsubsection{Training/Test Dataset Resolution Impact}\label{sect:results:deeponet:ttdri}

Now, we use the entire ground trust dataset (see Section \ref{sect:adapting:dataset}) for training and testing of a PI-DeepONet model with the multiple-input and multiple-output architecture proposed in Section \ref{sect:adapting:mydeeponet}, and the loss function given in Eq. \eqref{eq:adapting:pideeponet:loss}. 

\begin{figure}[ht]
	\centering	
	\resizebox{\columnwidth}{!}{\includegraphics[scale=0.5]{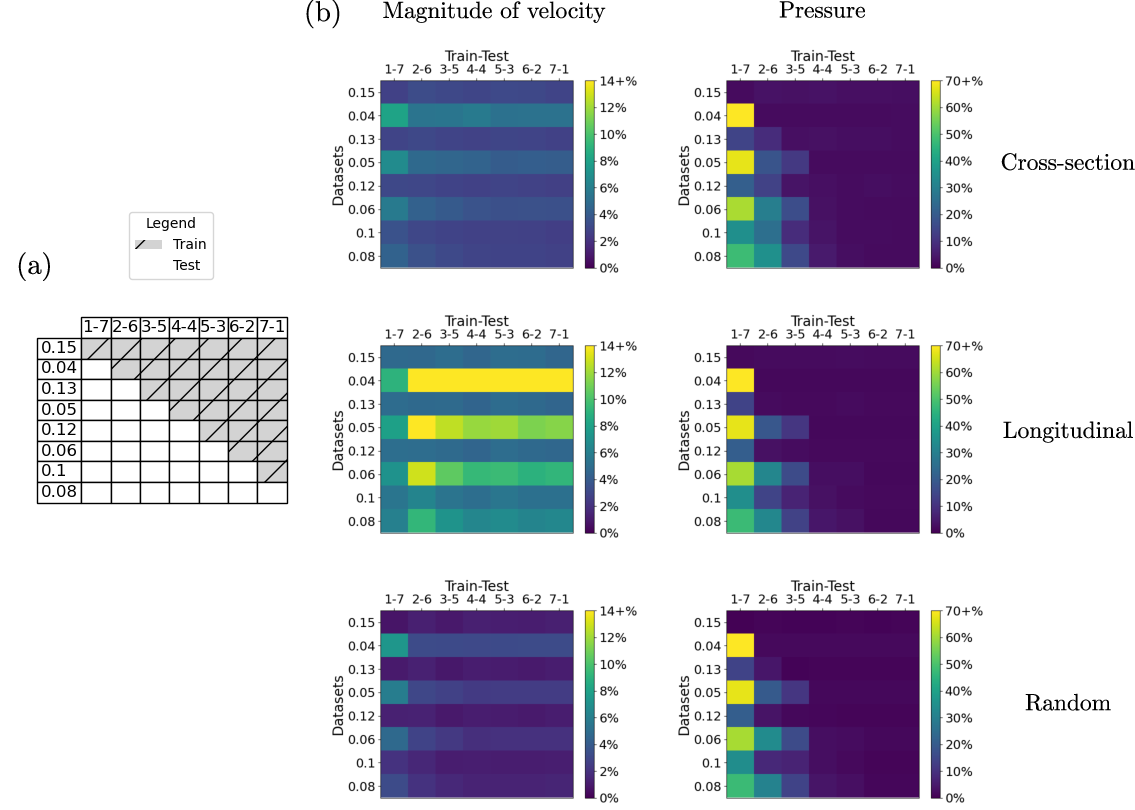}}
	\caption{Impact of training/test datasets split.}
	\label{fig:results:deeponet:datasetresolution}
\end{figure}

For this aim, we assess seven different scenarios in which we split the ground truth dataset into two groups, as illustrated in Figure \ref{fig:results:deeponet:datasetresolution}(a). For instance, the pair 3-5 means that datasets with $V\in [0.15,0.04,0.13]m/s$  are used for training the PI-DeepONet model, while datasets with $V\in [0.05,0.12,0.06,0.1,0.8]m/s$, not seen during training, are used for testing. The goal is to empirically determine the optimal split of the training and test datasets that maximizes the accuracy of the model when predicting both the magnitude of the velocity and the pressure. Note that the arrangement in Figure \ref{fig:results:deeponet:datasetresolution}(a) is purposely made so that in each scenario the training phase includes the maximum and minimum velocity values.

The architecture consists of three main components: branch1, branch2, and trunk (see Figure \ref{fig:adapting:mydeeponet}). Both branches use a MLP with 3 hidden layers and an input dimension of 1037 neurons, which corresponds to the number of mesh coordinate points in the inlet and outlet regions, respectively. Each hidden layer contains 100 neurons, and the output dimension is set to 400 neurons. The branches use the Tanh activation function and Xavier initialization to ensure balanced weight initialization. On the other hand, the trunk net uses a MLP with 4 hidden layers and 200 neurons each. The input dimension is 3, which corresponds to the coordinate points $(x_1, x_2, x_3)$, and the output dimension is also set to 400. Similarly to the branches, the trunk net uses the Tanh activation function and Xavier initialization. It is worth mentioning that from the total of 400 neurons in the output dimension, we will assign 100 neurons to each output variable, i.e., $\hat{\mathfrak{G}}^{(v_1)}_\theta$ (1-100), $\hat{\mathfrak{G}}^{(v_2)}_\theta$ (101-200), $\hat{\mathfrak{G}}^{(v_3)}_\theta$ (201-300), and $\hat{\mathfrak{G}}^{(p)}_\theta$ (301-400).

The PI-DeepONet model is trained for 200 000 iterations with a batch size of 1024, utilizing batch training and optimized with the Adam optimizer. In addition, we use the dimensionless form of NSE, the sampling within the mesh, and exponential decay for the learning rate combined with an optimizer scheduler.

Figure \ref{fig:results:deeponet:datasetresolution}(b) shows the percentage $L^2$-relative error on the training and test $V$-dependent datasets versus the considered training/test split scenario for the magnitude of the velocity and for the pressure. The results are also displayed across the three data configurations outlined in Figure \ref{fig:adapting:datascenarios}. The panels reveal an interesting pattern for both velocity and pressure predictions.

For velocity,  $V$-dependent datasets with lower maximum inlet velocities have the largest relative errors, indicating that they are more difficult for the model to learn. However, as the size of the datasets involved in the training phase increases, these errors tend to decrease, suggesting that more data helps to improve accuracy. The best overall performance is observed in the `random data' configuration, where predictions of the magnitude of the velocity are more accurate.

Regarding the pressure, starting from the scenario 3-5, all subsequent configurations show good prediction accuracy, indicating that the model becomes more reliable in predicting pressure as the size of the $V$-dependent datasets involved in the training phase increases.
\subsubsection{DeepONet Vs PI-DeepONet}\label{sect:results:deeponet:comp}
In this section, we compare DeepONet and PI-DeepONet approaches. The objective is to study the impact of incorporating the governing physical equations into the loss function of PI-DeepONet in contrast to the case of the DeepONet approach which is purely data-driven. The same three data formatting scenarios displayed in Figure \ref{fig:adapting:datascenarios} are considered. In addition, we use the multiple-input, multiple-output architecture proposed in Section \ref{sect:adapting:mydeeponet}, the loss functions given in Eqs. \eqref{eq:adapting:pideeponet:loss} and \eqref{eq:adapting:pideeponet:loss}, and the hyper-parameters and techniques as in Section \ref{sect:results:deeponet:ttdri}.

Based on the previous section results, we choose a scenario of type 5-3 for performing our comparison. In particular, the $V$-dependent datasets with $V\in[0.04, 0.06, 0.1, 0.12, 0.15] m/s$ are used for training the models, while the ones with $V\in[0.05, 0.08,0.13] m/s$ are used for testing. 

Table \ref{tab:results:deeponet:comp} presents the comparison results between the DeepONet and PI-DeepONet models, focusing on both performance and computational efficiency. In particular, we report the $L^2$-relative error on the training and test $V$-dependent datasets, which reflects the discrepancy between predicted and ground truth values for both velocity magnitude and pressure. We also provide the runtime required to train each model.

PI-DeepONet demonstrated higher accuracy than DeepONet in the first two cases, i.e., `cross-section' and `longitudinal', with the largest improvement observed in the `longitudinal' scenario. In the `random' case, the results of both models are more closely aligned. Overall, PI-DeepONet shows improved performance over DeepONet in terms of the magnitude of velocity, although it still faces challenges with $V$-dependent datasets with lower $V$ values. However, when it comes to pressure, DeepONet yields lower relative errors, indicating better performance in this study. In addition, training the DeepONet model is $1.6x$ faster in terms of computational efficiency compared to the PI-DeepOnet models, supporting the use of the former when appropriate.  

\begin{table}[ht]
	\centering
	\resizebox{16.5cm}{!}{	
		\begin{tabular}{||c|c|c||c|c|c||c|c|c||}
			\hline
			\multicolumn{1}{||c}{}  &\multicolumn{2}{|c||}{}  & \multicolumn{3}{|c||}{\textbf{DeepONet}}  & \multicolumn{3}{|c||}{\textbf{PI-DeepONet}}  \\
			\multicolumn{1}{||c}{ } & \multicolumn{2}{|c||}{ } & \multicolumn{2}{|c}{ $L^2$-relative error} & \multicolumn{1}{c||}{ }  & \multicolumn{2}{|c}{ $L^2$-relative error} & \multicolumn{1}{c||}{ }  \\ 
			\cline{4-9}
			Source of data  &  \multicolumn{2}{|c||}{\textbf{Datasets}}  &   \rotatebox{0}{Magnitude of velocity} &  \rotatebox{0}{Pressure}
			&  \rotatebox{0}{Runtime(h)} &  \rotatebox{0}{Magnitude of velocity} &  \rotatebox{0}{Pressure}
			&  \rotatebox{0}{Runtime(h)} \\ \hline
Cross-section / 5 slices & train &V004  &  1.2238e-01 & 8.4834e-02 & 8.93 & 5.4001e-02 & 2.0648e-02 & 14.52  \\
    &  &V006  &   1.1312e-01 &  8.9753e-02 &    & 3.4948e-02 & 2.0793e-02 & \\
    1.42\%  &  &V01  &  1.0223e-01 &  9.8372e-02 &  & 2.6794e-02 & 2.4334e-02 & \\
            &  &V012  &  9.8716e-02 &  1.0255e-01 &  & 2.7489e-02 & 2.7283e-02 & \\
            &  &V015  &  9.4953e-02 &  1.0834e-01 &  & 2.9623e-02 & 3.1953e-02 &  \\ \cline{2-6} \cline{7-9}
    &  test &V005  &  1.1727e-01 &  8.8884e-02 & & 4.2461e-02 & 2.3986e-02 & \\
            &  &V008  &  1.0679e-01 &  9.3711e-02 &  & 2.8150e-02 & 2.2159e-02&  \\
            & &V013  &  9.7288e-02 &  1.0455e-01 &  & 2.8119e-02 & 2.8774e-02 & \\
    \hline
Longitudinal / 1 slice & train &V004  &   1.9664e-01 &  4.8856e-01 & 8.93  &  1.4748e-01 &  1.5595e-02 & 14.66   \\
         &  & V006  &   1.9120e-01 &  5.1318e-01 &   &  9.2258e-02  &  1.4848e-02 &   \\
    1.87\%       &  & V01  &   1.8424e-01 &  5.6148e-01 &   &  4.9349e-02 &  1.5549e-02 &  \\
           &  & V012  &   1.8177e-01 &  5.8341e-01 &   &  4.3853e-02 &  1.6737e-02 &   \\
            &  & V015  &   1.7889e-01 &  6.1449e-01 &   &  4.3609e-02 &  1.9169e-02 &   \\ \cline{2-6} \cline{7-9}
             &  test  & V005  &   1.9376e-01 &  4.9920e-01 &   &  1.1667e-01 &  2.4341e-02 &   \\
              &  & V008  &   1.8726e-01 &  5.3853e-01 &   &  6.2953e-02 &  1.6272e-02
 &   \\
               &  & V013  &   1.8073e-01 &  5.9400e-01 &  &  4.2964e-02 &  1.7408e-02 &   \\
			\hline
Random / Coord. points & train &V004  &   1.7673e-02 &  6.2008e-03 & 8.93 & 3.1676e-02 & 1.7270e-02 & 14.74 \\
			&  & V006  &   1.7109e-02 &  6.4153e-03 &   &  2.0244e-02 &  1.4592e-02 &   \\
		0.3\% 	 &  & V01  &   1.7208e-02 &  7.5501e-03 &  & 1.1227e-02 & 9.9699e-03 &    \\
			 &  & V012  &   1.7494e-02 &  8.2986e-03 & & 1.0050e-02 & 7.7110e-03 &    \\
			 &  & V015  &   1.8183e-02 &  9.7196e-03 &  & 1.0552e-02 & 6.7369e-03 &     \\ \cline{2-6} \cline{7-9}
			 &  test  & V005  &  1.7228e-02 &  1.2525e-02 &  & 2.5097e-02 & 1.9196e-02 &   \\
			 &  & V008  &   1.7087e-02 &  7.7839e-03 &  & 1.4320e-02 & 1.5595e-02 &    \\
			 &  & V013  &   1.7694e-02 &  8.7354e-03 & & 9.9959e-03 & 7.0725e-03 &    \\
			\hline
	\end{tabular}} 
	\caption{Comparison between DeepONet and PI-DeepONet. }
    \label{tab:results:deeponet:comp}
\end{table}

It is worth mentioning that the runtime of the testing phase, or more specifically the inference in (PI-)DeepONets is highly efficient and faster than CFD simulations. For instance, in \cite{Junyan2024}, the authors highlighted that the inference time of the Sequential DeepONet (S-DeepONet) is at least three orders of magnitude faster than direct numerical simulations, demonstrating the model's efficiency in generating predictions. In our context, the CFD approach discussed in Section \ref{sect:cfd} takes approximately $3$ minutes per $V$-dependent dataset to compute the velocity and pressure fields for all mesh points. In contrast, the inference time for PI-DeepONets is significantly shorter, around $8$ seconds, making the process of inferring the velocity and pressure fields for a whole new $V$-dependent dataset $22.5\times$ faster.

\begin{figure}[ht]
	\centering	
    \includegraphics[scale=0.9]{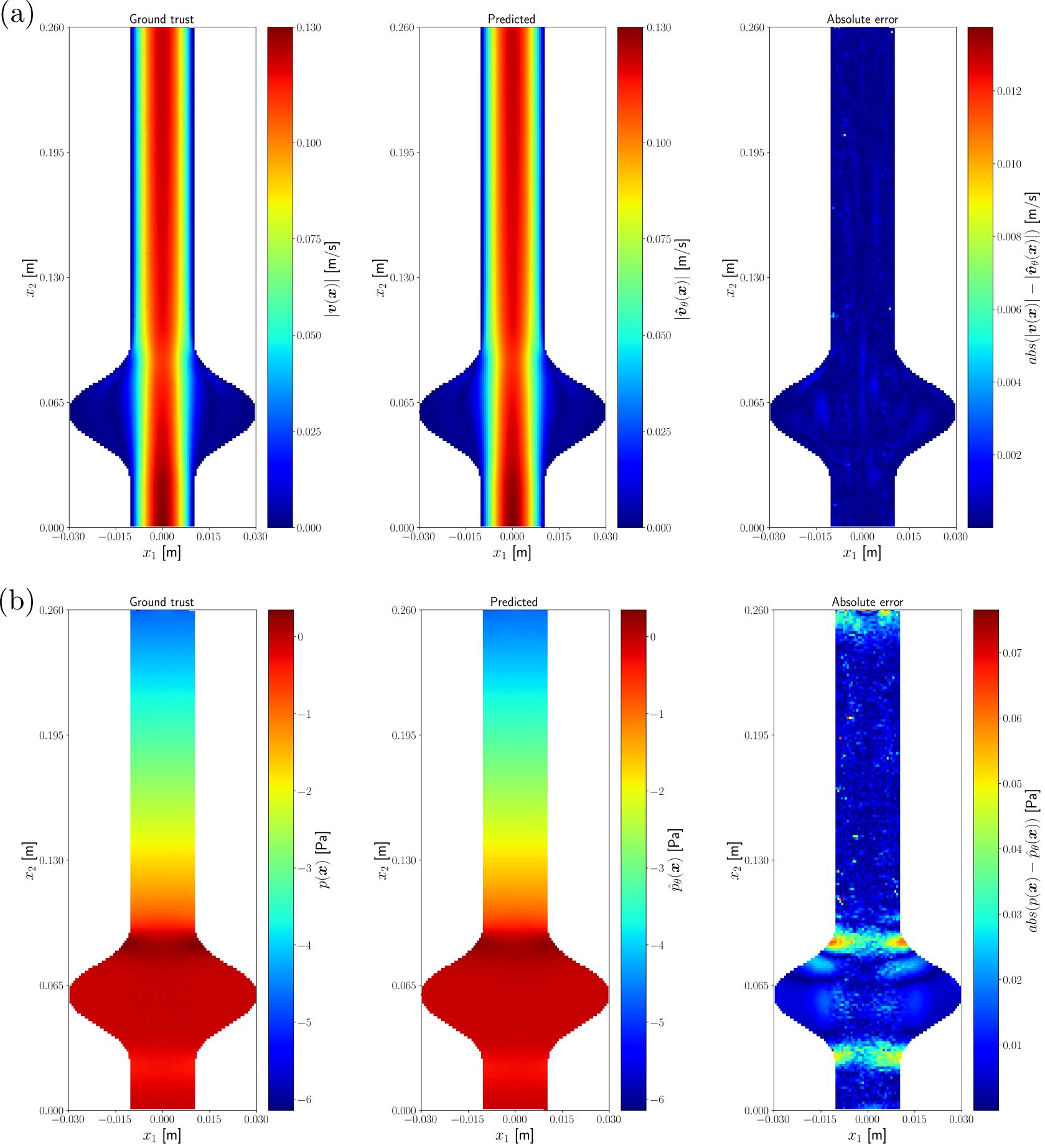}
	\caption{Comparison between the ground trust and the predicted values for (a) the magnitude of the velocity and (b) the pressure via absolute error along the plane $x_3=0$.}
	\label{fig:longitudinal}
\end{figure}

Finally, we retrieve from Table \ref{tab:results:deeponet:comp} the results using the dataset with $V=0.13\;m/s$ in the `random' scenario, and we show a comparison in Figure \ref{fig:longitudinal} between the ground trust and the predicted values of (a) the magnitude of the velocity and (b) the pressure field via a visual representation of the absolute error along the longitudinal plane $x_3=0$. The low observed absolute error evidences the capability of the applied method to accurately capture the velocity and pressure distributions, respectively. 

\section{Discussion and conclusions}\label{sect:conslusions}


This study has addressed the application of Physics-Informed Neural Networks (PINNs), Deep Operator Networks (DeepONets) and their Physics-Informed extensions (PI-DeepONets) in predicting steady vascular flow simulations in the context of a 3D Abdominal Aortic Aneurysm (AAA) idealized model. PINN and (PI-)DeepONet methods were adapted to suit this specific case, incorporating Navier-Stokes equations as physical laws governing fluid dynamics. PINN was applied in scenarios with and without Loss Data, following best practices to mitigate optimization pathologies. Regarding (PI-)DeepONet, we proposed an architecture that further enhances its capabilities, integrating multiple inputs and outputs while showing accurate results. For the sake of the validation, we compared against CFD simulations on benchmark datasets and demonstrated good agreement between the results.


In Section \ref{sect:results:pinns:drf}, we illustrated how PINNs can generalize predictions across an entire domain using sparse data, which may come from experimental measurements or numerical simulations, including cross-sectional, longitudinal, or randomly distributed data points. Moreover, PINNs are shown to be resilient to noisy data, often encountered in real-world applications (see Section \ref{sect:results:pinns:noise}). Additionally, PINNs provided robust initial predictions for flow simulations when the model was trained solely on governing equations and boundary conditions (see Section \ref{sect:results:pinns:pnd}). On accelerating the training runtime, in Section \ref{sect:results:pinns:drf}, we introduced WU-PINN, an approach that, framed in the Data scenario, allows to accelerate PINN by incorporating a preliminary DeepNN training phase. We also explored a transfer learning approach in Section \ref{sect:results:pinns:tl}, to enhance the training efficiency of PINNs, focusing on the data-less scenario.

One of the main limitations of PINNs is the training computation time, which represents here the bottleneck. Although the specific use case examined in this study does not present significant challenges to CFD simulations in terms of performance and computational efficiency, the runtime of the training phase shown by our implementation of the PINNs method is significantly higher than the computational time of CFD simulations. Even using transfer learning techniques.


On the other hand, (PI-)DeepONet offered key advantages when the maximum inlet velocity $V$ varied, i.e., the Parametric-PDEs scenario (see Sections \ref{sect:results:deeponet} and \ref{sect:results:deeponet:comp}). In contrast to PINNs, which need to be retrained for each new boundary condition, (PI-)DeepONets can manage varying conditions without requiring the time-consuming retraining process. This was performed by processing the training split of the entire ground trust dataset. In addition, the main strength lies in the ability to generalize inference to unseen input conditions, such as the possibility of inferring whole new $V$-dependent datasets, within a specified range of trained values of $V$. As shown in Section \ref{sect:results:deeponet:comp}, the inference runtime proved to be faster than the CFD simulations.

The proposed (PI-)DeepOnet methodology presents a significant potential, however, the main limitation is that it requires substantially more training data in comparison to PINNs, which can be costly to obtain. Our approach empirically demonstrated that even with a modest number of training instances (covering a finite range of maximum input velocity values), the (PI-)DeepONet model could generalize within that range. Nevertheless, the model may struggle to generalize to inputs outside the specified range of trained values.

In addition, the selective data usage during training (see Section \ref{sect:adapting:dataset}) provided a robust, generalizable approach. This was especially valuable for mimicking more complex scenarios where it is impractical to acquire complete data sets.


In terms of potential applications, our PINNs and (PI-)DeepONet approaches can be directly applied to a large variety of scenarios. For instance, to problems featuring similar steady flow configurations but varied geometries, as well as to scenarios involving different boundary conditions settings. 
  
In particular, the (PI-)DeepONet capacity for efficient generalization could be useful in medical applications requiring patient-specific solutions, where repeated evaluations are needed for individualized conditions. For instance, predicting patient-specific responses $u^{(i)}(\boldsymbol{x}, t)=\mathfrak{G}(f^{(i)})(\boldsymbol{x}, t)$ across various physiological inputs $f^{(i)}$ can be highly resource-intensive with traditional CFD solvers. However, for the inference, (PI-)DeepONet only performs a forward pass to generate predictions, which scales well with more complex inputs or larger networks. In addition, it can be further enhanced using hardware acceleration. So then, PI-DeepONet provides a fast, direct solution method, ideal for scenarios needing rapid or real-time results thanks to its powerful generalization, once trained.


 Further generalizations of our frameworks include their extensions to handle unsteady and transient flows, moving boundaries and eventually to capture fluid-structure interactions, and their adaptation to more complex biological structures. For unsteady flows, methods such as “Respecting Causality” \cite{Wang2024Causality} for PINNs, and "Long-time integration" \cite{Xu2023, Wang2023Long} or Sequential Deep Operator Network (\cite{He2024A}) for (PI-)DeepONet are available for enhancing the accuracy in predictions. Handling moving boundaries in these frameworks remains an open challenge, although recent works are beginning to emerge in this area, for example, \cite{Zhu2024} for PINNs and \cite{Xu2024} for DeepONet. In addition, parameterized PINNs (see \cite{Cho2024,Liu2024}) can be considered to address the retraining issue associated with changing conditions.

 Additionally, in our work, we ensured that the maximum input velocities $V$ were within a range in which the underlying physics did not change drastically. The study of an opposite scenario continues to be of major interest.
 
Future work could also be focused on automating the optimization of hyper-parameters using meta-learning techniques or hyper-parameter sweeps across learning rates, network architectures, and activation functions, as suggested in \cite{Chelsea2017} and \cite{Wang2023Long}. In addition, for a robust evaluation, it should be necessary to conduct multiple runs with varying random seeds.


\section*{Data and code availability}
The data and code that support the findings of this study are available from the corresponding author upon reasonable request.

\section*{CRediT authorship contribution statement}

\textbf{OLCG:} Conceptualization, Methodology, Software, Visualization, Formal analysis, Writing -- original draft. \textbf{BG:} Funding
acquisition, Conceptualization, Methodology, Formal analysis, Writing -- review \& editing. \textbf{VD:} Funding acquisition, Conceptualization, Methodology, Formal analysis, Writing -- review \& editing.

\section*{Declaration of competing interest}

The authors declare that they have no known competing financial interests or personal relationships that could have
appeared to influence the work reported in this paper.

 \section*{Acknowledgements}\label{Acknowledgements}
This work received support from French government under the France 2030 investment plan, as a part of the Initiative d'Excellence d'Aix-Marseille Université - A*MIDEX, AMX-21-RID-022.

\appendix

\newpage
\section{Dimensions of a training dataset in DeepONet} \label{appendix:deeponet:dataset:dim}

Consider a single sample where $N=1$ and $P=1$, thus the dataset triplet becomes,
\begin{align}
	&\left[
	\left[
	\begin{array}{c}
		\boldsymbol{x}_1^{(1)}, t_1^{(1)}
	\end{array}
	\right]; 
	\quad
	\left[
	\begin{array}{ccc}
	f^{(1)}(\tilde{\boldsymbol{x}}_1), f^{(1)}(\tilde{\boldsymbol{x}}_2), \cdots, f^{(1)}(\tilde{\boldsymbol{x}}_m)
	\end{array}
	\right]; 
	\quad
	\left[
	\begin{array}{c}
	\mathfrak{G}(f^{(1)}) ( \boldsymbol{x}_1^{(1)}, t_1^{(1)} )
	\end{array}
	\right]
	\right].
	\end{align}

Next, if we select only one input function $f^{(1)}$ and $P$ different points from the domain of the target solution function $u$, labeled as $(\boldsymbol{x}_1^{(1)}, t_1^{(1)}), (\boldsymbol{x}_2^{(1)}, t_2^{(1)}), \dots, (\boldsymbol{x}_P^{(1)}, t_P^{(1)})$ i.e., $N=1, P > 1$, $f^{(1)}$ repeats itself for $P$ times, and the triplet dataset now is given by,
\begin{align}
&\left[
\left[
\begin{array}{c}
	\boldsymbol{x}_1^{(1)}, t_1^{(1)} \\
	\boldsymbol{x}_2^{(1)}, t_2^{(1)} \\
\vdots \\
\boldsymbol{x}_P^{(1)}, t_P^{(1)} \\
\end{array}
\right]; 
\quad
\left[
\begin{array}{ccc}
f^{(1)}(\tilde{\boldsymbol{x}}_1), f^{(1)}(\tilde{\boldsymbol{x}}_2), \cdots, f^{(1)}(\tilde{\boldsymbol{x}}_m) \\
f^{(1)}(\tilde{\boldsymbol{x}}_1), f^{(1)}(\tilde{\boldsymbol{x}}_2), \cdots, f^{(1)}(\tilde{\boldsymbol{x}}_m) \\
\vdots \\
f^{(1)}(\tilde{\boldsymbol{x}}_1), f^{(1)}(\tilde{\boldsymbol{x}}_2), \cdots, f^{(1)}(\tilde{\boldsymbol{x}}_m) \\
\end{array}
\right]; 
\quad
\left[
\begin{array}{c}
\mathfrak{G}(f^{(1)}) ( \boldsymbol{x}_1^{(1)}, t_1^{(1)} )\\
\mathfrak{G}(f^{(1)}) ( \boldsymbol{x}_2^{(1)}, t_2^{(1)} ) \\
\vdots \\
\mathfrak{G}(f^{(1)}) ( \boldsymbol{x}_P^{(1)}, t_P^{(1)} ) \\
\end{array}
\right]
\right].
\end{align}

Finally, the most general scenario arises when $N$ sample functions are also considered, i.e., $N > 1, P > 1$. Here, for each sample $f^{(i)}$, $\mathfrak{G}(f^{(i)})$ is evaluated at P different points from the domain $( \boldsymbol{x}_j^{(i)}, t_j^{(i)} )$, and the triplet dataset becomes,
\begin{align}
	& \begin{bmatrix}
	\left[
	\begin{array}{c}
		\boldsymbol{x}_1^{(1)}, t_1^{(1)} \\
		\boldsymbol{x}_2^{(1)}, t_2^{(1)} \\
	\vdots \\
	\boldsymbol{x}_P^{(1)}, t_P^{(1)} \\
	\end{array}
	\right]; 
	\quad
	\left[
	\begin{array}{ccc}
	f^{(1)}(\tilde{\boldsymbol{x}}_1), f^{(1)}(\tilde{\boldsymbol{x}}_2), \cdots, f^{(1)}(\tilde{\boldsymbol{x}}_m) \\
	f^{(1)}(\tilde{\boldsymbol{x}}_1), f^{(1)}(\tilde{\boldsymbol{x}}_2), \cdots, f^{(1)}(\tilde{\boldsymbol{x}}_m) \\
	\vdots \\
	f^{(1)}(\tilde{\boldsymbol{x}}_1), f^{(1)}(\tilde{\boldsymbol{x}}_2), \cdots, f^{(1)}(\tilde{\boldsymbol{x}}_m) \\
	\end{array}
	\right]; 
	\quad
	\left[
	\begin{array}{c}
	\mathfrak{G}(f^{(1)}) ( \boldsymbol{x}_1^{(1)}, t_1^{(1)} )\\
	\mathfrak{G}(f^{(1)}) ( \boldsymbol{x}_2^{(1)}, t_2^{(1)} ) \\
	\vdots \\
	\mathfrak{G}(f^{(1)}) ( \boldsymbol{x}_P^{(1)}, t_P^{(1)} ) \\
	\end{array}
	\right]\\\\
	\left[
	\begin{array}{c}
		\boldsymbol{x}_1^{(2)}, t_1^{(2)} \\
		\boldsymbol{x}_2^{(2)}, t_2^{(1)} \\
	\vdots \\
	\boldsymbol{x}_P^{(2)}, t_P^{(2)} \\
	\end{array}
	\right]; 
	\quad
	\left[
	\begin{array}{ccc}
	f^{(2)}(\tilde{\boldsymbol{x}}_1), f^{(2)}(\tilde{\boldsymbol{x}}_2), \cdots, f^{(2)}(\tilde{\boldsymbol{x}}_m) \\
	f^{(2)}(\tilde{\boldsymbol{x}}_1), f^{(2)}(\tilde{\boldsymbol{x}}_2), \cdots, f^{(2)}(\tilde{\boldsymbol{x}}_m) \\
	\vdots \\
	f^{(2)}(\tilde{\boldsymbol{x}}_1), f^{(2)}(\tilde{\boldsymbol{x}}_2), \cdots, f^{(2)}(\tilde{\boldsymbol{x}}_m) \\
	\end{array}
	\right]; 
	\quad
	\left[
	\begin{array}{c}
	\mathfrak{G}(f^{(2)}) ( \boldsymbol{x}_1^{(2)}, t_1^{(2)} )\\
	\mathfrak{G}(f^{(2)}) ( \boldsymbol{x}_2^{(2)}, t_2^{(2)} ) \\
	\vdots \\
	\mathfrak{G}(f^{(2)}) ( \boldsymbol{x}_P^{(2)}, t_P^{(2)} ) \\
	\end{array}
	\right] \\\\
		\begin{array}{c}
		\vdots \\
		\end{array} 
		\quad\quad\quad\quad\quad\quad\quad\quad\quad\quad\quad\quad\quad
		\begin{array}{ccc}
		\vdots \\
		\end{array}
		\quad\quad\quad\quad\quad\quad\quad\quad\quad\quad\quad\quad
		\begin{array}{c}
		\vdots \\
		\end{array} \quad\\\\
	\left[
	\begin{array}{c}
		\boldsymbol{x}_1^{(N)}, t_1^{(N)} \\
		\boldsymbol{x}_2^{(N)}, t_2^{(N)} \\
	\vdots \\
	\boldsymbol{x}_P^{(N)}, t_P^{(N)} \\
	\end{array}
	\right]; 
	\quad
	\left[
	\begin{array}{ccc}
	f^{(N)}(\tilde{\boldsymbol{x}}_1), f^{(N)}(\tilde{\boldsymbol{x}}_2), \cdots, f^{(N)}(\tilde{\boldsymbol{x}}_m) \\
	f^{(N)}(\tilde{\boldsymbol{x}}_1), f^{(N)}(\tilde{\boldsymbol{x}}_2), \cdots, f^{(N)}(\tilde{\boldsymbol{x}}_m) \\
	\vdots \\
	f^{(N)}(\tilde{\boldsymbol{x}}_1), f^{(N)}(\tilde{\boldsymbol{x}}_2), \cdots, f^{(N)}(\tilde{\boldsymbol{x}}_m) \\
	\end{array}
	\right]; 
	\quad
	\left[
	\begin{array}{c}
	\mathfrak{G}(f^{(N)}) ( \boldsymbol{x}_1^{(N)}, t_1^{(N)} )\\
	\mathfrak{G}(f^{(N)}) ( \boldsymbol{x}_2^{(N)}, t_2^{(N)} ) \\
	\vdots \\
	\mathfrak{G}(f^{(N)}) ( \boldsymbol{x}_P^{(N)}, t_P^{(N)} ) \\
	\end{array}
	\right]
\end{bmatrix}
\end{align}

\newpage
\section{Dimensionless form of NSE for AAA idealized model} \label{appendix::dimensionless}
In this section, we derive the dimensionless form of Eqs. \eqref{eq:formulation:nse:momentum3D}-\eqref{eq:formulation:nse:bc:outlet}. For this purpose, we choose constant parameters that are representatives of the physical problem we are trying to solve. In particular, we use ($V$) and ($D$), representing the maximum inlet velocity and vessel diameter at the inlet, respectively. So then, we normalize the physical quantities as follows,
\begin{align}
    & \boldsymbol{x}^{*} = \frac{\boldsymbol{x}}{D}, \quad \boldsymbol{v}^{*} = \frac{\boldsymbol{v}}{V}, \quad \boldsymbol{p}^{*} = \frac{\boldsymbol{p}}{\rho_f V^2}. \label{quantities:normalization}
\end{align}

Notice that, from Eq. \eqref{quantities:normalization} we can derive that,
\begin{align}
    & \nabla^{*} = D \nabla, \quad (\nabla^{*})^2 = D^2 \nabla^2. \label{nabla:normalization}
\end{align}

Based on Eqs. \eqref{quantities:normalization}-\eqref{nabla:normalization}, we rewrite Eqs. \eqref{eq:formulation:nse:momentum3D}-\eqref{eq:formulation:nse:bc:outlet} in a dimensionless form as follows,
\begin{subequations} 
\begin{align}
    & \frac{\rho_f V^2}{D} (\boldsymbol{v}^* \cdot \nabla^*)\boldsymbol{v}^* =  -\frac{\rho_f V^2}{D}\nabla^* p^* + \frac{V\mu_f}{D^2} (\nabla^*)^2 \boldsymbol{v}^*, \label{equations:normalization:a}\\
    &   \nabla^* \cdot \boldsymbol{v}^* = 0. \label{equations:normalization:b} 
\end{align}
\end{subequations}

At this point, if we multiply Eqs. \eqref{equations:normalization:a}-\eqref{equations:normalization:b} by $\displaystyle{\frac{D}{\rho_f V^2}}$ on both sides, we obtain
\begin{subequations} 
\begin{align}
    & (\boldsymbol{v}^* \cdot \nabla^*)\boldsymbol{v}^* =  -\nabla^* p^* + \frac{\mu_f}{\rho_f D V} (\nabla^*)^2 \boldsymbol{v}^*, \\
    &  \nabla^* \cdot \boldsymbol{v}^* = 0. 
\end{align}
\end{subequations}

Thus, the Reynolds number appears in the form $\displaystyle{Re = \frac{\rho_f D V}{\mu_f}}$, leading to the dimensionless form of NSE,
\begin{subequations} 
\begin{align}
    &  (\boldsymbol{v}^* \cdot \nabla^*)\boldsymbol{v}^* =  -\nabla^* p^* + \frac{1}{Re} (\nabla^*)^2 \boldsymbol{v}^*, \\
    &  \nabla^* \cdot \boldsymbol{v}^* = 0.
\end{align}
\end{subequations}
\newpage
\bibliographystyle{elsarticle}
\bibliography{Bibliography}


\end{document}